\documentclass{article} 
\usepackage{main,times}

\usepackage{amsmath,amsfonts,bm}

\def\eqref#1{equation~\ref{#1}}

\def\1{\bm{1}}

\DeclareMathAlphabet{\mathsfit}{\encodingdefault}{\sfdefault}{m}{sl}
\SetMathAlphabet{\mathsfit}{bold}{\encodingdefault}{\sfdefault}{bx}{n}

\usepackage{hyperref}
\usepackage{url}
\usepackage{enumitem}
\usepackage{graphicx}
\usepackage{booktabs}
\usepackage{algorithm}
\usepackage{algpseudocode}
\usepackage{multirow}

\usepackage{booktabs}
\usepackage[table]{xcolor}

\usepackage{subcaption}

\definecolor{ourrow}{HTML}{D6E4F7}   
\definecolor{panelbg}{HTML}{D9D9D9}    
\newcommand{\std}[1]{\,{\scriptsize\color{black!45}$\pm$#1}}
\newcommand{\best}[1]{\textbf{#1}}                 
\newcommand{\sbest}[1]{\underline{#1}}           
\newcommand{\up}{{\scriptsize\color{black!55}$\uparrow$}}
\newcommand{\down}{{\scriptsize\color{black!55}$\downarrow$}}

\usepackage{tabularray}
\UseTblrLibrary{booktabs}

\title{Giving Credit Where It’s Due: Redundancy-Aware Learning for Efficient Reasoning}

\author{
Yuqing Zhou\textsuperscript{1}\thanks{Work done during the internship at Amazon.},
Hong Wang\textsuperscript{2},
Manqing Mao\textsuperscript{2},
Zhuoer Wang\textsuperscript{2},
Samson Koelle\textsuperscript{2}, \\
\textbf{Jie Yuan\textsuperscript{2},
Yanjun Lin\textsuperscript{2},
James Feng\textsuperscript{2},
Nikki Lijing Kuang\textsuperscript{2},
Ziwei Zhu\textsuperscript{1},
Wei Niu\textsuperscript{2}}\thanks{Corresponding author.} \\
\textsuperscript{1}George Mason University,
\textsuperscript{2}Amazon, Inc \\
\texttt{\{yzhou31,zzhu20\}@gmu.edu}, \texttt{niuwei@amazon.com}
}

\iclrfinalcopy 
\begin{document}

\maketitle

\lhead{Preprint.}

\begin{abstract}
Large reasoning models can produce correct yet unnecessarily long reasoning traces. Existing methods improve reasoning efficiency with trajectory-level objectives or local token- and step-level signals, but rarely model inter-step semantic dependencies. This limits their ability to distinguish redundant steps from those that support later deductions, making it harder to shorten reasoning without sacrificing accuracy.
We introduce \textbf{RECAP} (\underline{RE}dundancy-aware \underline{C}redit \underline{A}ssignment via \underline{P}ropagation), which addresses this limitation by \textbf{assigning credit where it is due} 
based on both a step's downstream role in the reasoning structure and its contribution to solving the problem correctly.
We define \emph{structural responsibility} to capture the step’s downstream role by measuring how strongly later reasoning depends on it, using credit propagated backward from the final-answer node through an outcome-independent, LLM-annotated semantic dependency graph. \textbf{However, a step can have high structural responsibility yet steer the reasoning away from the correct solution.} RECAP therefore introduces \emph{step efficacy} to measure answer-directed progress through changes in gold-answer log-likelihood as each step is added. Together, these signals reshape rollout-level GRPO advantages into step-specific updates. RECAP requires neither a separately trained process reward model nor preconstructed concise trajectories. Across two 7B models and four mathematical reasoning benchmarks, RECAP improves the accuracy–efficiency trade-off. On Qwen2.5-Math-7B, it improves pass@1 by 2.0-3.7 percentage points while reducing reasoning tokens by $8\%–31\%$ relative to GRPO across all four benchmarks. Analysis suggests these savings reflect fewer reasoning operations and less dead-end reasoning, rather than more compact expression.
\end{abstract}

\section{Introduction}
Large reasoning models achieve strong performance on challenging mathematical and logical tasks by 
learning to generate extended reasoning traces through reinforcement learning with verifiable rewards~\citep{guo2025deepseek, jaech2024openai}. Although extended reasoning improves performance, not every step within a reasoning trace contributes equally to the final answer. Even successful traces can contain repetition, unproductive exploration, and unnecessary verification that contribute little to the final solution~\citep{pmlr-v267-chen25bx,sui2025stop,yue2025don}. These low-contribution steps introduce unnecessary computation and, in turn, increase inference-time cost. Standard outcome-based reinforcement learning provides no direct pressure to reduce this overhead, as it rewards a trace based on final-answer correctness~\citep{shao2024deepseekmath}. The resulting trajectory-level advantage is applied across the generated sequence, allowing low-contribution steps to be reinforced alongside the deductions that support the answer and thereby permitting redundant reasoning to persist. 

Recent work improves reasoning efficiency by assigning efficiency signals at different granularities. Trajectory-level methods encourage shorter solutions using explicit length signals~\citep{NEURIPS2025_579b5b84,li-etal-2026-leash,peng2026think,li2026drpo} or model-derived objectives such as entropy~\citep{huang2026pear}, but assign a single signal to the entire trace and therefore cannot identify which steps should be preserved. Finer-grained methods estimate token- or step-level utility using proxies such as attention or changes in correct-answer likelihood~\citep{liu-etal-2026-tokens, he2026iapo, li2026stepwise}. These signals provide more selective supervision, but primarily capture a step's direct relation to the final answer and may miss contributions realized only through later deductions. Pruning methods similarly inherit this limitation when removing low-utility segments or training on compressed traces~\citep{liu2026thoughtfold}. Structure-aware methods introduce trees, graphs, or information-flow signals to organize reasoning or assign finer-grained credit~\citep{ji2026tree,dong2026how,zhan2026graphpo}, but structural dependence alone does not reveal whether a step advances the correct solution.
This exposes a fundamental limitation of existing credit signals: \textbf{step contribution cannot be determined locally, but depends on how information propagates through later reasoning and toward the correct solution.}

To address this limitation, we propose \emph{REdundancy-aware Credit Assignment via Propagation (RECAP)}, which assigns step-level credit along two complementary dimensions: \textbf{whether later reasoning depends on the step}, and \textbf{whether that contribution advances the correct solution}. To capture downstream contribution, RECAP represents each trajectory as a typed semantic dependency graph with three relations: \emph{support}, \emph{context}, and \emph{restatement}, distinguishing information that is required for later deductions, provides useful context, or merely repeats what is already available. RECAP then propagates credit backward from the final answer through this graph, assigning greater \textbf{structural responsibility} to steps whose information supports downstream reasoning and less to redundant or unused steps. Structural responsibility alone, however, does not indicate whether a step advances the correct solution: \emph{an incorrect step may still support an entire downstream reasoning.} RECAP therefore complements structural responsibility with \textbf{step efficacy}, measured by the change in correct-answer log-likelihood as each step is added. We couple efficacy with the sign of the rollout-level advantage, reinforcing answer-advancing steps on successful rollouts while protecting them from excessive penalty on unsuccessful ones.

RECAP combines step efficacy with the propagated structural responsibility into a step-specific weight, which refines the outcome-derived advantage into an \emph{outcome-aware, dependency-sensitive} training signal for each step. It requires neither a separately trained process reward model nor preconstructed concise trajectories. In summary, our contributions are:
\begin{enumerate}[label=\textbf{\arabic*.}, leftmargin=*, nosep]
    \item We introduce a typed semantic dependency graph that distinguishes three inter-step relations: \emph{support}, \emph{context}, and \emph{restatement}. We propagate structural credit through these dependencies, assigning each step credit by its downstream role rather than its distance from the final answer.
    \item We complement structural credit with a step-level efficacy signal that measures each step's progress toward the correct answer. Combining the two signals turns the rollout-level GRPO advantage into step-specific updates that reinforce useful reasoning while downweighting redundancy in successful trajectories, and preserve helpful steps while penalizing misleading ones in unsuccessful trajectories.
    \item Extensive experiments across models and mathematical benchmarks show that RECAP maintains or improves accuracy while reducing reasoning tokens, cutting them by $8\%–31\%$ on Qwen2.5-Math-7B versus GRPO while improving pass@1 on all four benchmarks, primarily by reducing unnecessary computation such as dead-end reasoning. 
\end{enumerate}

\section{Related Work}
Existing approaches to token-efficient reasoning can be broadly grouped into prompt-based control, length-regularized policy optimization, pruning and distillation, and fine-grained credit assignment. Prompt-based methods encourage concise reasoning through explicit instructions, output constraints, or adaptive hints~\citep{10852493,lee2025how,tang2025concisehint}, but provide limited guidance on which reasoning steps should be preserved.
\paragraph{Length Regularization and Trace Compression.}
Length-regularized methods incorporate response length into rollout-level optimization through fixed or adaptive penalties conditioned on correctness, problem difficulty, or training dynamics~\citep{NEURIPS2025_579b5b84,li2026drpo,peng2026think,li-etal-2026-leash}. These methods can effectively shorten responses, but their efficiency signal is defined at the trajectory level and does not identify which individual steps are redundant. A complementary line of work constructs concise training traces by pruning redundant content or distilling compressed trajectories~\citep{jiang-etal-2026-drp,liu-etal-2026-pru,ma-etal-2026-reasoning,yuan-etal-2026-graph-based,liu2026thoughtfold}. Such methods learn from compressed traces rather than assigning credit to steps within newly sampled rollouts.
\paragraph{Fine-Grained and Structure-Aware Credit Assignment.}
More recent methods assign token- or step-level signals using answer likelihood, confidence, conditional mutual information, or attention~\citep{li-etal-2026-think-better,wang-etal-2026-stabilizing,he2026iapo,nie-etal-2026-attnpo,liu-etal-2026-tokens,li2026stepwise}. These signals provide more selective supervision, but largely evaluate each unit through its local effect and do not explicitly capture how information introduced by one step contributes to later reasoning. Structure-aware methods introduce trees or graphs to organize reasoning trajectories or assign finer-grained credit~\citep{ji2026tree,dong2026how,zhan2026graphpo}. However, they either target more efficient sampling and credit assignment rather than shorter reasoning traces, or improve token efficiency through shorter paths without directly identifying redundant computation.

RECAP directly models these intra-trace semantic dependencies and propagates structural responsibility through the resulting graph, while using step efficacy to incorporate answer-directed progress. This allows policy updates to distinguish steps by both their downstream role and their effect on the final solution. We provide a more detailed comparison in Appendix~\ref{app:related-work}.

\section{Preliminary}
RECAP extends Group Relative Policy Optimization (GRPO)~\citep{shao2024deepseekmath} with step-specific credit assignment. For each prompt, GRPO samples a group of rollouts and assigns each rollout an outcome reward. Given reward $R_k$ for rollout $k$, GRPO computes the group-normalized advantage $A_k^{GRPO}=\frac{R_k - mean(\mathbf{R})}{std(\mathbf{R}) + \epsilon}$, 
where $\epsilon>0$ is used for numerical stability. Rollouts with above-average rewards receive positive advantages, while those below the group average receive negative advantages. Then, it assigns $A_{k,t}=A_k^{GRPO}$ for all tokens $t$ within the rollout and optimize
\begin{equation}
J(\theta)=\mathbb{E}\left[\frac{1}{G}\sum_{k=1}^{G}\frac{1}{|o_k|}\sum_{t=1}^{|o_k|}\left(\min\left(r_{k,t}A_{k,t},\;\mathrm{clip}(r_{k,t},1-\varepsilon,1+\varepsilon)A_{k,t}\right)-\beta_{DL}\,\mathbb{D}_{KL}\right)\right]
\label{eq:optimize_obj}
\end{equation}
The expectation is over queries $q$ sampled from the distribution $Q$, and rollouts $o_{k}\sim\pi_{\theta_{old}}(\cdot\mid q)$. $\pi_{\theta}$ and $\pi_{\theta_{old}}$ denote the current and old policy models, respectively. $G$ is the number of rollouts in each group. The token-level importance sampling ratio is $r_{k,t}=\frac{\pi_\theta(o_{k,t}|q,o_{k,<t})}{\pi_{\theta_{old}}(o_{k,t}|q,o_{k,<t})}$.
Clipping this ratio limits drastic policy updates, while $\beta_{DL}\mathbb{D}_{KL}$ is a $KL$ divergence term that regularizes the current policy $\pi_{\theta}$ against large deviation from the reference policy $\pi_{ref}$.

A key limitation of GRPO lies in the uniform assignment $A_{k,t}=A_k^{\mathrm{GRPO}}$. All tokens in a rollout receive the same learning signal despite contributing differently to the final solution. Useful derivations, repeated checks, and abandoned branches are therefore reinforced or penalized together. To reduce redundant reasoning, we seek to redistribute the rollout-level signal $A_k^{\mathrm{GRPO}}$ across the reasoning trace according to the contribution of each step. Intuitively, a step should receive greater credit when its information is subsequently used by later reasoning and contributes to the final answer. This leads to our central question: \emph{how should learning credit be distributed across reasoning steps so that, on the same problems, the learned policy preserves successful reasoning while suppressing redundant content and using fewer inference tokens?}

\section{Methodology}
\label{sec:method}
The question of how to assign the advantage at the step-level cannot be judged from the step alone. A locally correct step may still be redundant if nothing depends on it, while a short transition may be essential if it supports later deductions. We therefore model dependencies among reasoning steps explicitly and use them to construct dependency-aware step credit. As shown in Figure~\ref{fig:overview}, the main design of RECAP has three stages:
\begin{enumerate}[label=\textbf{\arabic*.}, leftmargin=*, nosep]
    \item \textbf{Structural Responsibility within a trace.} We build a semantic dependency graph within each rollout to identify which steps are used by later reasoning or the final answer (Section~\ref{subsec:dependency-model}), and propagate structure credit backward from the final answer (Section~\ref{subsec:structure-credit-prop}).
    \item \textbf{Step Efficacy toward the solution.} Besides the structual responsibilty, we measure whether each step moves the reasoning toward or away from the correct answer (Section~\ref{sec:step-efficacy}).
    \item \textbf{Reshape the rollout-level advantage.} We combine structural responsibility and step efficacy to reshape the rollout-level GRPO advantage into step-specific advantages (Section~\ref{sec:adv-shaping}).
\end{enumerate}

\subsection{Structural Responsibility within a Trace}
\label{sec:structure-credit}

A model could waste tokens by restating existing information or introducing content unused by later reasoning. Therefore, improving efficiency requires distinguishing steps that carry useful information forward from those that do not. We characterize this distinction through a step's \emph{structural responsibility}, which measures how strongly subsequent deductions depend on the information it introduces. Specifically, we (1) construct an intra-trace semantic dependency graph to represent how information is carried forward, and (2) propagate credit over the graph to quantify each step's responsibility for the downstream reasoning leading to the final answer.

\subsubsection{Intra-Trace Dependency Modeling}
\label{subsec:dependency-model}
A step's contribution depends on how later reasoning uses it, so we represent each trace as a semantic dependency directed acyclic graph (DAG) rather than a plain sequence. We segment each rollout into steps as nodes, and an edge $i\to j$ means that Step $j$ uses information from Step $i$. Dependencies point forward in the trace, thus the graph is acyclic. Steps with no directed path to the final answer form disconnected branches.

\begin{figure}[t]
\begin{center}
\includegraphics[width=0.9\textwidth]{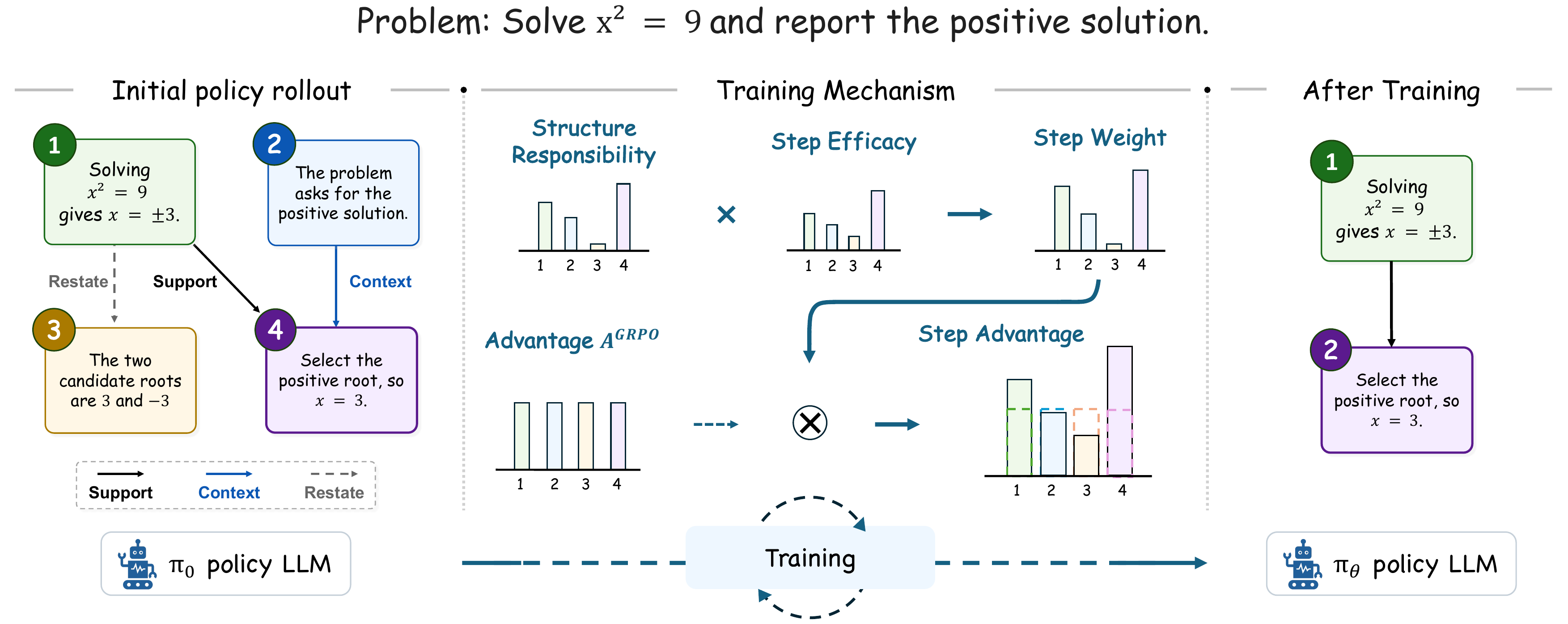}
\end{center}
\vspace{-5pt}
\caption{The overview of RECAP.}
\label{fig:overview}
\end{figure}

Not every dependency carries the same weight. A step may be strictly necessary for subsequent steps, merely helpful, or entirely useless. Collapsing these cases into a single edge type would obscure these distinctions. Therefore, we define three edge types and use LLM-as-a-Judge~\citep{gu2026survey} to annotate pairwise step dependencies:
\begin{itemize}[leftmargin=*]
    \item \textit{\textbf{Support}}. Step j directly requires Step i for its derivation, verification, or answer extraction. These edges form the backbone of the solution. 
    \item \textit{\textbf{Context}}. Step i supplies useful framing for Step j but is not logically required, which is helpful, yet removable without breaking the derivation. 
    \item \textit{\textbf{Restate}}. Step j repeats what Step i already established without making reasoning progress, which is a marker of redundancy rather than contribution. 
\end{itemize}
Figure~\ref{fig:overview} shows an example of the three edge types and how we turn a trace into a graph with them.

\subsubsection{Dependency-based Structure Credit Propagation}
\label{subsec:structure-credit-prop}
To leverage the structural contributions captured by the graph for policy optimization, we convert it into a structure responsibility score for each step, quantifying its contribution to the final answer. We propagate responsibility backward under a local conservation rule. Specifically, we initialize the final-answer node with a responsibility $resp(F)=1$, so that \textbf{the responsibility is outcome-independent}, and each node then distribute its responsibility back to the prerequisite steps on which it depends. For an edge $s_{k,i}\to s_{k,j}$ between step $i$ and step $j$ of rollout $k$, each child $s_{k,j}$ distributes the responsibility among its parents $s_{k,j}$ in proportion to their edge coefficients, normalized over $s_{k,j}$'s incoming edges. This gives the following update for each step $s_{k,i}$: 
\begin{equation}
    resp(s_{k,i}) = \sum_{j\in\{j'|\exists i\to j'\}} \frac{\gamma_{i\to j}}{W_j}resp(s_{k,j}),\qquad W_j=\sum_{p'\in\mathrm{parents}(j)}\gamma_{p'\to j}.  
\end{equation}
The coefficient $\gamma_{p\to c}$ is determined by the edge type, so responsibility flows according to the role of each dependency: support edges carry the most, context edges partial, and restate edges none. Normalizing over each child's incoming edges ensures it distributes only the responsibility it holds across its parents, preventing an ``in-degree hack" in which adding more dependencies could otherwise duplicate responsibility. The resulting responsibility of a step reflects how much structural responsibility reaches it through downstream dependencies. Backbone steps that support more downstream reasoning receive more, while redundant branches accumulate little. Because step responsibility follows semantic dependencies rather than positional distance, essential prerequisites can retain high responsibility regardless of how early they appear.

\subsection{Step Efficacy toward the Solution}
\label{sec:step-efficacy}
Structural responsibility measures how strongly downstream reasoning depends on a step, but not whether the step moves the model toward the correct solution. Two steps may have similar structural responsibility while having opposite effects on the correctness of the solution. We therefore introduce \emph{step efficacy} to measure each step's immediate progress toward or away from the correct answer, complementing structural responsibility with an answer-directed signal.

Let the gold answer for the prompt $q$ be tokenized as $a^\star = (a^\star_1,\ldots a^\star_T)$. For rollout $\tau_k$, let $\tau_{k, \leq i} = (s_{k,1}, \ldots,s_{k,i})$ denote the trajectory prefix ending at step $s_{k,i}$. Using teacher forcing, we evaluate the gold answer under each prefix by computing its mean token log-likelihood:
\begin{equation}
    L_{k,i}=\frac{1}{T}\sum^T_{t=1}\log\pi_\theta (a^\star_t\mid q,\tau_{k,\leq i}, a^\star_{<t}),
\label{eq:prefix_gold_prob}
\end{equation}
where $T$ is the number of the gold answer tokens and $a^\star_{<t}$ denotes the gold-answer tokens preceding $a^\star_{t}$. For $i=0$, we compute $L_{k,0}$ from the prompt $x$ without any generated steps. We define the efficacy of step $s_{k,i}$ as the change in the mean gold-answer log-likelihood caused by extending the preceding trajectory prefix with that step:
\begin{equation}
    \Delta_{k,i} = L_{k,i} - L_{k,i-1}
\label{eq:step_efficacy}
\end{equation}
A positive $\Delta_{k,i}$ indicates that the step moves the model toward the gold answer, while a negative value indicates the opposite. Step efficacy therefore captures the step's immediate answer-directed effect under the current policy.

\subsection{From step scores to policy-gradient advantages}
\label{sec:adv-shaping}

Structural responsibility captures downstream dependence, while step efficacy measures whether a step moves the model toward or away from the gold answer. We combine these complementary signals into a multiplicative step score that jointly captures downstream dependence and answer-directed effect, using an exponential transformation to convert efficacy into a positive factor on structural responsibility. Since exponentiation can amplify extreme efficacy values, we first clip the efficacy as $\widehat{\Delta}_{k,i}=\operatorname{clip}(\Delta_{k,i},-c,c)$, where $c>0$ is a fixed threshold.
The effective contribution score of step $s_{k,i}$ is then
\begin{equation}
  M_{k, i} = resp(s_{k,i}) * e^{\alpha * \operatorname{sign}(A_k)\widehat{\Delta}_{k,i} }
\end{equation}
where $\alpha > 0$ controls the influence of step efficacy and $\operatorname{sign}(A_k)$ denotes the sign of the rollout-level advantage. The sign term makes the efficacy adjustment consistent with the rollout-level learning direction. Positive efficacy increases the step score in positive-advantage rollouts but decreases it in negative-advantage rollouts, while negative efficacy has the opposite effect. Thus, answer-advancing steps are reinforced more strongly in positive-advantage rollouts and protected from excessive penalty in negative-advantage rollouts. Structural responsibility independently modulates these updates according to each step's downstream use.

The scale of step scores may vary across rollouts, so we normalize them within each trace before reshaping the policy-gradient advantage. We compute the token-weighted mean $\bar{M}_{k} = \sum_i \frac{d_{k,i}}{D_k} M_{k,i}$, where $d_{k,i}$ is the number of tokens in step $i$ and $D_k=\sum_i d_{k,i}$, and normalize each step score as $w_{k,i} =  \frac{M_{k,i}}{ \bar{M}_{k} + \epsilon}$, where $\epsilon >0$ ensures numerical stability. This normalization expresses each step score relative to the token-weighted mean within the rollout, preserving the rollout-level update scale while redistributing credit across steps.

Finally, we use $w_{k,i}$ to adjust the rollout-level GRPO advantage. Specifically, instead of assigning a uniform advantage to the entire rollout, RECAP assigns all tokens within step $i$ with the advantage
\begin{equation}
    A^{RECAP}_{k,i} =  [(1-\beta) + \beta * w_{k,i}] A_{k}.
\end{equation}
where $\beta\in [0,1]$ controls the strength of step-level credit assignment.
Since $w_{k,i}$ is normalized so that its expectation over tokens within each rollout is approximately one, the expected RECAP advantage over tokens $t$ satisfies $\mathbb{E}_{t\sim\tau_k}\left[A_{k,t}^{\mathrm{RECAP}}\right]\approx A_k^{\mathrm{GRPO}}$.
RECAP therefore redistributes the original GRPO advantage across the trace while approximately preserving its overall scale and update direction. Ideally, in positive-advantage rollouts, structurally important, answer-advancing steps have above-average effective contribution and thereby receive stronger updates, while redundant, weakly connected, or counterproductive steps are downweighted. In negative-advantage rollouts, answer-advancing steps are protected from excessive penalty, whereas structurally important, answer-degrading steps receive stronger penalties. The resulting RECAP advantages replace the original GRPO advantages in Eq.~\ref{eq:optimize_obj}, and the complete procedure is summarized in Algorithm~\ref{alg:recap}.

\section{Experiments}
\label{sec:experiments}
Our evaluation addresses three research questions (RQs): RECAP's accuracy--efficiency trade-off (\textbf{RQ1}), the contributions of structural responsibility and step efficacy (\textbf{RQ2}), and the source of its token savings (\textbf{RQ3}). We first describe the experimental setup, then address these questions in turn.
\paragraph{Datasets.} 
We randomly sample 2048 problems from DAPO-Math-17k~\citep{yu2025dapoopensourcellmreinforcement} for training and evaluate on four mathmatical reasoning benchmarks of increasing difficulty: GSM8K~\citep{cobbe2021gsm8k} for simple grade school problems, MATH-500~\citep{lightman2024let} for more challenging competition mathematics, and AIME 2024~\citep{aime24} and AIME 2025~\citep{aime25} for highly challenging competition problems. 

\paragraph{Models.} 
To test whether RECAP generalizes across different reasoning regimes, we evaluate two representative 7B models: Qwen2.5-Math-7B~\citep{yang2024qwen25mathtechnicalreportmathematical}, a non-thinking model, and DeepSeek-R1-Distill-Qwen-7B~\citep{deepseekai2025deepseekr1incentivizingreasoningcapability}, a reasoning model.
\paragraph{Baselines.} 
We restrict baselines to standard autoregressive reasoning methods for directly comparable token efficiency and exclude structure-aware methods whose primary objectives are sampling or attribution during training rather than inference efficiency, and discuss them separately in Related Work.
We compare RECAP against eight baselines: the base model; a prompt-based baseline that encourages concise step-by-step reasoning~\citep{10852493}; standard GRPO~\citep{shao2024deepseekmath}; GRPO with a simple length penalty; DDCA~\citep{peng2026think}, which restricts length optimization to correct rollouts and adapts its strength by problem difficulty; LEASH~\citep{li-etal-2026-leash}, which dynamically adjusts a length penalty to enforce a target response length; BINGO~\citep{liu-etal-2026-tokens}, which applies stronger efficiency pressure to less important tokens; and ThoughtFold~\citep{liu2026thoughtfold}, which prunes redundant rollout segments, constructs preference pairs across compression levels, and combines preference optimization with GRPO. 

\paragraph{Training \& Evaluation Details.}
We use Claude Haiku 4.5~\citep{anthropic2025claudethaiku45} for dependency annotation.  The annotation is outcome-independent. The annotator sees neither the gold answer nor solution trajectories, and only labels dependencies among steps. Annotator selection and annotation quality are evaluated in Appendix~\ref{app:annotation_quality}. For training, we sample $16$ rollouts per prompt with temperature $1.0$ and a maximum response length of $8192$ tokens. We train on 2K prompts for four epochs ($60$ steps) with batch size $128$, mini-batch size $1024$, two policy-update epochs, and AdamW at a constant learning rate of $1\times10^{-6}$. RECAP modifies only advantage construction and all other training settings are shared with the baselines. We select checkpoints by greedy $pass@1$ on a held-out 512-prompt DAPO-Math split evaluated every $15$ steps. Full details are provided in Appendix~\ref{app:training-details}. For evaluation, following DeepSeek-R1~\citep{deepseekai2025deepseekr1incentivizingreasoningcapability}, we sample $k=8$ rollouts per prompt with temperature $0.6$ and top-$p=0.95$, and report $pass@1$, $pass@k$, and mean generated tokens.

\subsection{Overall Accuracy--Efficiency Trade-off (RQ1)}
\begin{figure}[ht]
    \centering
    \begin{subfigure}[t]{0.45\textwidth}
        \centering
        \includegraphics[width=\linewidth]{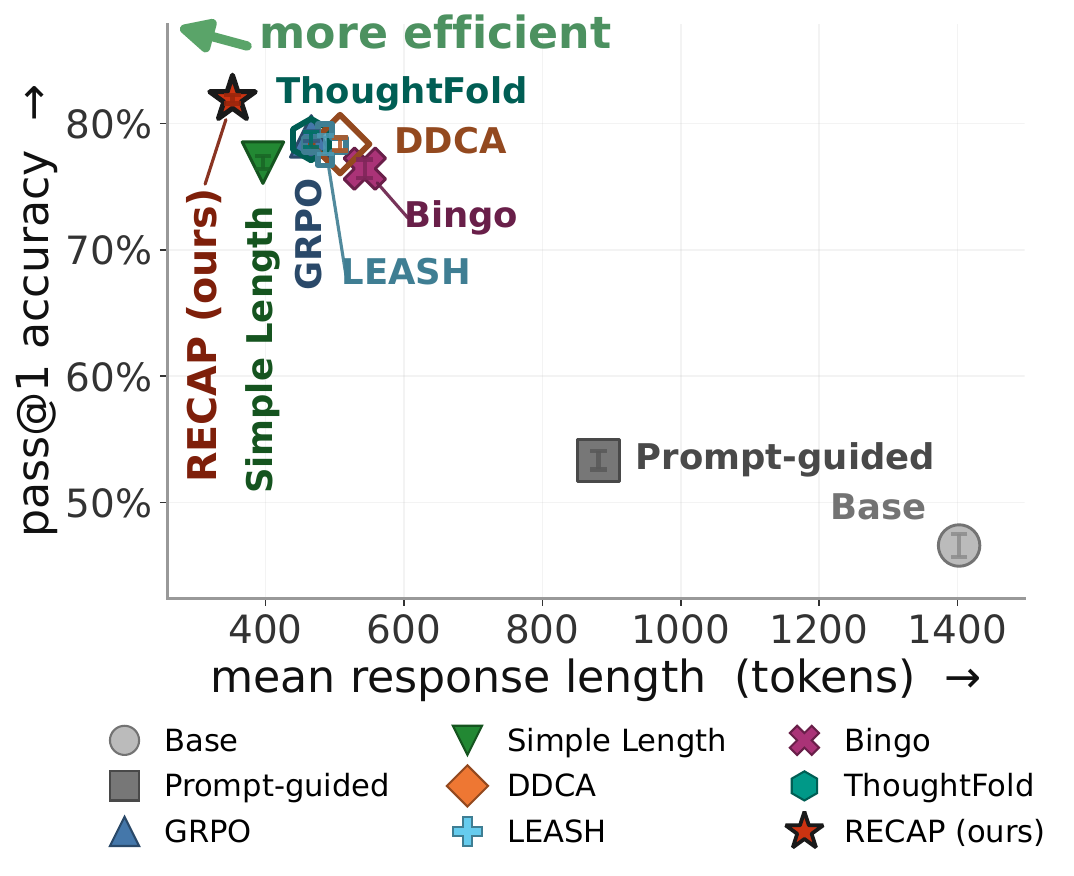}
        \caption{Qwen2.5-Math-7B}
        \label{fig:qwen25}
    \end{subfigure}
    \hfill
    \begin{subfigure}[t]{0.45\textwidth}
        \centering
        \includegraphics[width=\linewidth]{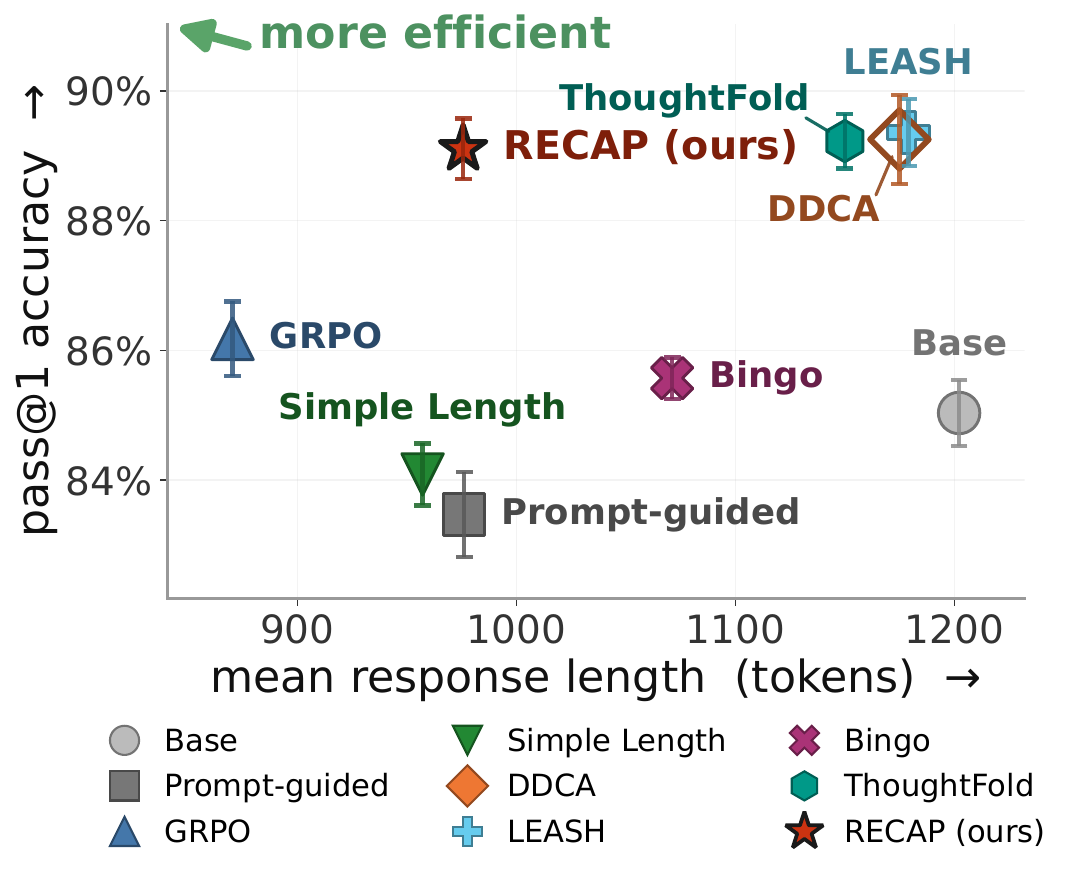}
        \caption{DeepSeek-R1-Distill-Qwen-7B}
        \label{fig:deepseek}
    \end{subfigure}
    \caption{Accuracy--efficiency trade-off across models.}
    \label{fig:main_results}
\end{figure}

We first examine whether RECAP improves reasoning efficiency without sacrificing task performance. Figure~\ref{fig:main_results} summarizes the average accuracy--efficiency trade-off across the four benchmarks and Table~\ref{tab:main} reports the results across the benchmarks.

\paragraph{RECAP improves both accuracy and efficiency on Qwen2.5-Math-7B.} As shown in Figure~\ref{fig:main_results}, RECAP lies on the upper-left frontier, achieving a favorable reduction in reasoning cost without sacrificing accuracy. Table~\ref{tab:main} shows that this advantage is consistent across tasks. It achieves the highest $pass@1$ on all four benchmarks, while producing the shortest responses on GSM8K and MATH-500 and the second-shortest responses on AIME 2024 and AIME 2025. Compared with GRPO, RECAP reduces response length by $31\%$, $18\%$, $8\%$, and $22\%$ on GSM8K, MATH-500, AIME 2024, and AIME 2025, respectively, while improving $pass@1$ by $3.3$, $2.0$, $2.5$, and $3.7$ percentage points. These results show a consistent shift toward higher accuracy at lower inference cost, rather than shorter reasoning obtained by sacrificing solution quality.

\paragraph{RECAP retains high accuracy at substantially lower cost on DeepSeek-R1-Distill-Qwen-7B.}
RECAP remains in the highest-accuracy region while using substantially fewer tokens than methods with comparable accuracy.  
With averages weighted by the number of evaluation examples in each benchmark, RECAP saves at least $170$ tokens per rollout compared with each of DDCA, LEASH, and ThoughtFold.
On the three harder benchmarks, it produces the second-shortest responses, behind only GRPO, while remaining close to the best $pass@1$. 
GRPO is shorter, but its $pass@1$ is lower by $2.0$--$5.4$ percentage points across the four benchmarks. Overall, RECAP achieves the most favorable accuracy--efficiency trade-off among the compared methods.

\begin{table*}[t]
\centering
\small
\caption{Sampled $pass@1$ (\%) and mean completion tokens across four benchmarks. Grey $\pm$ values are the standard deviations over $8$ runs. \textbf{Bold}/\underline{underline} mark best and second best per block.}
\label{tab:main}
\begin{tblr}{
  colspec   = {lcccccccc},
  colsep    = 4pt,
  rowsep    = 0.8pt,
  column{1} = {leftsep  = 0pt},
  column{9} = {rightsep = 0pt},
}
\toprule
\SetCell[r=2]{l,m} Method & \SetCell[c=2]{c} GSM8K & & \SetCell[c=2]{c} MATH-500 & & \SetCell[c=2]{c} AIME 2024 & & \SetCell[c=2]{c} AIME 2025 & \\
\cmidrule[leftpos=0,rightpos=0]{2-3} \cmidrule[leftpos=0,rightpos=0]{4-5} \cmidrule[leftpos=0,rightpos=0]{6-7} \cmidrule[leftpos=0,rightpos=0]{8-9}
 & pass@1\up & tokens\down & pass@1\up & tokens\down & pass@1\up & tokens\down & pass@1\up & tokens\down \\
\midrule
\SetRow{bg=panelbg, abovesep=2pt, belowsep=2pt} \SetCell[c=9]{c} \textit{Qwen2.5-Math-7B} & & & & & & & & \\
Base & 49.1\std{1.2} & 1253\std{101} & 44.6\std{1.1} & 1684\std{75} & 11.2\std{5.6} & 2464\std{308} & \phantom{0}4.2\std{2.4} & 2206\std{736} \\
\SetRow{belowsep=0pt} Prompt-guided & 57.0\std{0.8} & \phantom{0}746\std{62} & 49.3\std{1.7} & 1121\std{74} & 10.4\std{5.5} & 1892\std{362} & \phantom{0}2.1\std{2.5} & 1799\std{228} \\
\hline[0.05em]
\SetRow{belowsep=0pt} GRPO & \sbest{84.4}\std{0.4} & \phantom{0}354\std{6} & 72.2\std{1.0} & \phantom{0}641\std{24} & 24.2\std{3.9} & 1529\std{389} & \phantom{0}8.8\std{2.5} & 1405\std{262} \\
\hline[0.05em]
Simple Length & 82.2\std{0.5} & \phantom{0}\sbest{308}\std{3} & 70.2\std{1.3} & \phantom{0}\sbest{534}\std{40} & \sbest{26.2}\std{5.8} & \best{1333}\std{440} & \phantom{0}6.2\std{4.9} & \best{1037}\std{251} \\
DDCA & 83.9\std{0.6} & \phantom{0}380\std{4} & 71.5\std{1.0} & \phantom{0}702\std{34} & 20.4\std{7.0} & 1883\std{218} & \phantom{0}7.9\std{1.7} & 1503\std{334} \\
LEASH & 83.4\std{0.9} & \phantom{0}373\std{4} & 72.1\std{1.2} & \phantom{0}654\std{28} & \sbest{26.2}\std{7.2} & 1738\std{363} & \sbest{11.2}\std{3.5} & 1400\std{312} \\
Bingo & 82.2\std{0.9} & \phantom{0}400\std{9} & 68.6\std{1.2} & \phantom{0}785\std{30} & 21.2\std{6.7} & 1789\std{513} & \phantom{0}8.3\std{4.4} & 1571\std{275} \\
\SetRow{belowsep=0pt} ThoughtFold & 84.0\std{0.6} & \phantom{0}364\std{6} & \sbest{72.3}\std{1.4} & \phantom{0}617\std{16} & 22.1\std{5.9} & 1404\std{271} & \sbest{11.2}\std{1.7} & 1469\std{174} \\
\SetRow{bg=ourrow} \textbf{RECAP (ours)} & \best{87.7}\std{0.3} & \phantom{0}\best{246}\std{3} & \best{74.2}\std{1.0} & \phantom{0}\best{524}\std{16} & \best{26.7}\std{3.1} & \sbest{1401}\std{211} & \best{12.5}\std{1.5} & \sbest{1098}\std{182} \\
\bottomrule
\SetRow{bg=panelbg, abovesep=2pt, belowsep=2pt} \SetCell[c=9]{c} \textit{DeepSeek-R1-Distill-Qwen-7B} & & & & & & & & \\
Base & 87.4\std{0.6} & \phantom{0}463\std{4} & 84.7\std{1.0} & 2488\std{53} & 43.8\std{5.2} & 6628\std{226} & 27.9\std{3.5} & 6837\std{148} \\
\SetRow{belowsep=0pt} Prompt-guided & 87.2\std{0.8} & \phantom{0}\best{430}\std{2} & 79.9\std{1.2} & 1758\std{26} & 33.8\std{4.2} & 6204\std{153} & 28.7\std{4.3} & 6726\std{180} \\
\midrule
\SetRow{belowsep=0pt} GRPO & 90.1\std{0.5} & \phantom{0}518\std{4} & 82.3\std{1.6} & \best{1347}\std{23} & 39.2\std{8.1} & \best{4856}\std{279} & 25.4\std{6.4} & \best{4432}\std{365} \\
\midrule
Simple Length & 87.7\std{0.6} & \phantom{0}\sbest{447}\std{2} & 80.4\std{0.8} & 1681\std{19} & 41.2\std{3.5} & 5828\std{233} & 29.6\std{3.3} & 6449\std{190} \\
DDCA & 91.9\std{0.9} & \phantom{0}670\std{9} & \sbest{88.3}\std{0.9} & 1951\std{35} & \best{45.8}\std{7.1} & 5644\std{216} & \best{32.1}\std{4.3} & 5964\std{236} \\
LEASH & \sbest{92.0}\std{0.6} & \phantom{0}666\std{12} & \best{88.5}\std{1.0} & 1983\std{42} & \sbest{45.0}\std{6.9} & 5666\std{308} & \best{32.1}\std{4.7} & 5840\std{188} \\
Bingo & 87.7\std{0.4} & \phantom{0}459\std{3} & 85.7\std{0.5} & 2061\std{34} & 44.2\std{4.3} & 6208\std{255} & \underline{31.3}\std{1.7} & 6344\std{155} \\
\SetRow{belowsep=0pt} ThoughtFold & \best{92.1}\std{0.5} & \phantom{0}657\std{5} & 87.9\std{0.7} & 1900\std{42} & 44.6\std{7.1} & 5802\std{274} & 29.6\std{4.9} & 5666\std{148} \\
\SetRow{bg=ourrow} \textbf{RECAP (ours)} & \best{92.1}\std{0.6} & \phantom{0}599\std{5} & 87.4\std{0.6} & \sbest{1513}\std{29} & 44.2\std{6.4} & \sbest{4895}\std{378} & 30.8\std{5.0} & \sbest{4659}\std{297} \\
\bottomrule
\end{tblr}
\end{table*}

\subsection{Contributions of Structural Responsibility and Step Efficacy (RQ2)}
\label{sec:signal-ablation}

\begin{table*}[ht]
\centering
\caption{\textbf{Ablation of the two signals in RECAP's step score} on Qwen2.5-Math-7B across four benchmarks. All runs share $\beta=0.3$, $\gamma_{context}=0.5$, $16$ rollouts per prompt and $60$ steps.}
\label{tab:ablation-signals}
\begin{tabular}{@{}lccc@{}}
\toprule

       & pass@1$\uparrow$ & pass@8$\uparrow$ & tokens$\downarrow$\\
\midrule
\textbf{RECAP (full)} & \textbf{81.90} & 91.80 & 352 \\
\midrule
w/o structural responsibility & 78.30 & 91.75 & 459 \\
w/o step efficacy  & 81.32 & \textbf{92.18} & 370 \\
Shuffled structural responsibility & 79.10 & 91.86  & \textbf{340} \\
Random structural responsibility & 78.45 & 92.12 & 343  \\
\bottomrule
\end{tabular}
\vspace{-5pt}
\end{table*}

We examine the contributions of RECAP's two signals by ablating either structural responsibility or step efficacy. To test whether structural responsibility benefits from its semantic alignment rather than merely non-uniform weighting, we additionally replace it with shuffled or random credit. Shuffling preserves the responsibility-score distribution within each rollout while breaking its correspondence to reasoning steps. Table~\ref{tab:ablation-signals} reports the results.

\paragraph{Structural responsibility drives the accuracy--efficiency gain.}
Removing structural responsibility decreases $pass@1$ from $81.90$ to $78.30$ while increasing the average response length from $352$ to $459$ tokens. The simultaneous degradation in accuracy and efficiency indicates that step efficacy alone cannot reliably identify the reasoning steps that should receive stronger updates. Structural responsibility is essential for concentrating updates on consequential steps, allowing RECAP to generate shorter reasoning traces without sacrificing accuracy.

\paragraph{Step efficacy further calibrates structural credit.}
Removing step efficacy reduces $pass@1$ by only $0.58$ points but increases mean response length by $\sim5\%$. Structural credit captures downstream dependence but not whether a step advances the correct solution. Step efficacy adds an answer-directed calibration, making credit assignment more selective and reducing unnecessary reasoning.

\paragraph{Semantic dependency enables effective credit assignment.}
Replacing structural responsibility with shuffled or random credit slightly shortens the average response, but reduces $pass@1$ by $2.80$ and $3.45$ points, respectively. The shuffled variant preserves the responsibility-score distribution while breaking its alignment with reasoning steps, showing that semantic alignment is critical for preserving accuracy rather than merely inducing non-uniform reweighting. Random credit provides complementary evidence that arbitrary non-uniform weighting is insufficient. These results demonstrate the importance of semantic dependency modeling for effective credit assignment.

Overall, structural responsibility is the main source of RECAP's accuracy--efficiency improvement, while step efficacy acts as a complementary calibration signal. Their combination yields the strongest $pass@1$ performance at nearly the shortest response length.

\subsection{Where Do RECAP's Token Savings Come From? (RQ3)}
\label{sec:rq3}
We next investigate whether RECAP saves tokens by expressing the same reasoning more compactly or by removing unnecessary computation. We measure how the number and token cost of reasoning units change during training, using equations and sentences as automatic proxies and manually annotating atomic reasoning operations on a subset of rollouts. Atomic operations are minimal reasoning steps that change the solution state or introduce new information used in the derivation, including setup, theorem recall, equation derivation, substitution, numerical computation, case analysis, verification, and conclusion. Detailed annotation criteria are provided in Appendix~\ref{app:atomic_op_def}.

\begin{figure}
    \centering
    \begin{subfigure}[t]{0.4\textwidth}
        \centering
        \includegraphics[width=\linewidth]{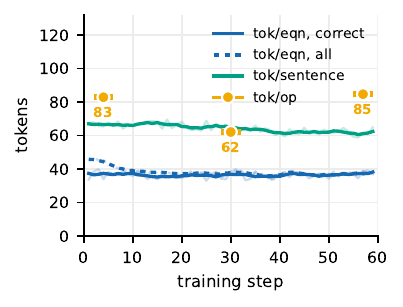}
        \caption{Token cost per reasoning operation.}
        \label{fig:A}
    \end{subfigure}
    \begin{subfigure}[t]{0.4\textwidth}
        \centering
        \includegraphics[width=\linewidth]{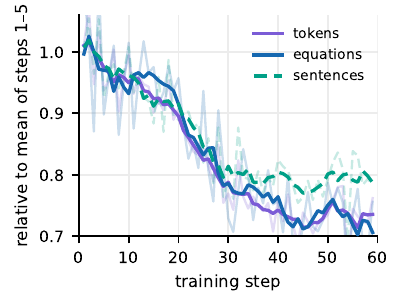}
        \caption{Evolution of reasoning unit counts.}
        \label{fig:B}
    \end{subfigure}
   
    \caption{\textbf{Training dynamics of reasoning content and token usage.} 
    (a) Tokens per reasoning unit, measured by equations, sentences, and manually annotated atomic operations. (b) Evolution of response tokens, equations, and sentences per rollout, normalized to their early-training values.
    }
    \label{fig:training}
\vspace{-2pt}
\end{figure}

\vspace{-10pt}
\paragraph{RECAP performs fewer reasoning operations rather than compressing them.}
Figure~\ref{fig:A} shows that the token cost per reasoning unit remains largely stable during training. On correct rollouts, tokens per equation change only from $36.4$ to $36.8$, with manually annotated operations showing the same pattern. In contrast, Figure~\ref{fig:B} shows that total tokens fall by $23.7\%$, alongside reductions of $24.5\%$ in equations and $17.3\%$ in sentences. Together, these results suggest that RECAP shortens responses primarily by performing fewer reasoning operations rather than expressing the same operations with fewer tokens.

\vspace{-10pt}
\paragraph{RECAP substantially reduces dead-end reasoning.}
We next measure the fraction of semantic steps with no directed path to the final answer, reported in Table~\ref{tab:deadend}. Across three independent annotators, RECAP consistently has the lowest dead-end rate. Only $5.4$--$5.7\%$ of its steps are dead ends, compared with $9.6$--$13.6\%$ for Simple Length and $19.5$--$25.3\%$ for GRPO. The same pattern holds when weighting by inference tokens, with only $3.3$--$4.0\%$ of RECAP's tokens spent on dead-end steps, versus $6.4$--$10.4\%$ for Simple Length and $11.1$--$17.2\%$ for GRPO. RECAP also yields the lowest or tied-lowest incidence of isolated steps across annotators. Complementary topology metrics in Appendix Table~\ref{tab:dag-topology} show the same pattern, with RECAP reducing both multiple-sink DAGs (ML) and isolated steps (ISO) across all three annotators.

Taken together, these results suggest that RECAP's efficiency gains reflect a structural simplification of reasoning. The model performs fewer reasoning operations, with a smaller fraction devoted to dead-end branches, rather than merely expressing the same computation with fewer tokens.

\begin{table*}[t]
\centering\small\setlength{\tabcolsep}{3.5pt}

\caption{\textbf{Dead-end reasoning in Qwen2.5-Math-7B test rollouts} (200 sampled prompts each from GSM8K and MATH500, all AIME prompts, and $8$ rollouts per prompt for each method). A step is a \emph{dead end} if it has no directed path to the final-answer node $F$, and \emph{isolated} if its total degree is zero. \textbf{step \%} and \textbf{tok \%} are macro-averaged per-rollout fractions by step count and token count, respectively. All three annotators label the same rollouts under the same segmentation. All standard errors are at most $0.6$\,pp and are omitted. Best (lowest) value per column in \textbf{bold}.}
\label{tab:deadend}
\small
\begin{tabular}{@{}lccccccccccccc@{}}
\toprule
& \multicolumn{6}{c}{Dead end (no path to $F$)} & \multicolumn{6}{c}{Isolated (degree $0$)} & \\
\cmidrule(lr){2-7}\cmidrule(lr){8-13}
& \multicolumn{3}{c}{step \% $\downarrow$} & \multicolumn{3}{c}{tok \% $\downarrow$} & \multicolumn{3}{c}{step \% $\downarrow$} & \multicolumn{3}{c}{tok \% $\downarrow$} & mean \\
\cmidrule(lr){2-4}\cmidrule(lr){5-7}\cmidrule(lr){8-10}\cmidrule(lr){11-13}
Method & Haiku & Opus & GPT & Haiku & Opus & GPT & Haiku & Opus & GPT & Haiku & Opus & GPT & steps \\
\midrule
Base & 26.4 & 29.1 & 23.7 & 22.9 & 25.8 & 21.6 & 7.6 & 15.3 & 10.6 & 6.8 & 13.8 & 9.7 & 12.69 \\
Prompt-guided & 25.0 & 25.4 & 21.4 & 21.8 & 22.2 & 19.6 & 8.8 & 13.2 & 10.2 & 8.0 & 12.0 & 9.4 & 11.21 \\
\midrule
GRPO & 25.3 & 19.5 & 20.1 & 17.2 & 11.1 & 12.6 & 10.0 & 11.6 & 13.4 & 4.0 & 4.7 & 7.0 & 9.25 \\
Simple Length & 13.6 & 9.6 & 9.7 & 10.4 & 6.4 & 7.3 & 2.8 & 4.0 & 5.4 & 1.4 & 2.2 & 3.5 & 7.49 \\
DDCA & 20.7 & 12.1 & 13.1 & 16.7 & 8.4 & 10.0 & 1.9 & 3.9 & 5.1 & 1.0 & 1.6 & 3.0 & 10.62 \\
LEASH & 19.7 & 12.3 & 12.7 & 15.7 & 8.2 & 9.7 & \textbf{1.6} & 3.4 & 4.9 & \textbf{0.8} & 1.2 & 2.9 & 10.04 \\
Bingo & 22.0 & 15.4 & 14.5 & 17.0 & 10.6 & 11.3 & 3.1 & 4.4 & 5.3 & 1.7 & 2.0 & 3.4 & 10.96 \\
ThoughtFold & 17.8 & 14.3 & 14.5 & 12.8 & 8.7 & 10.2 & 1.7 & 4.0 & 5.9 & \textbf{0.8} & 1.3 & 3.3 & 9.39 \\
\textbf{RECAP (ours)} & \textbf{5.6} & \textbf{5.7} & \textbf{5.4} & \textbf{3.9} & \textbf{3.3} & \textbf{4.0} & \textbf{1.6} & \textbf{1.7} & \textbf{3.5} & \textbf{0.8} & \textbf{0.8} & \textbf{2.1} & \textbf{4.29} \\
\bottomrule
\end{tabular}
\end{table*}

\section{Conclusion}
We introduced RECAP, a redundancy-aware credit assignment method that uses semantic dependencies and step-level reasoning progress to reshape GRPO updates for more efficient reasoning. Across two 7B models and four mathematical reasoning benchmarks, RECAP improves the accuracy--efficiency trade-off over trajectory-level, compression-based, and fine-grained baselines. Our analysis suggests that these gains come from reducing unnecessary reasoning computation rather than merely expressing the same reasoning more compactly, with substantially less inference budget spent on dead-end reasoning.

\subsection*{AI use statement}
Generative AI tools were used to assist with coding and debugging, including writing scripts for data analysis and plotting figures, literature search for related work, and refining the manuscript. All AI-assisted outputs were reviewed by the authors. The authors take the responsibility for the final content of this work.

\subsection*{Ethics statement}
This work studies reinforcement learning methods for improving the efficiency of large language model reasoning.  We use publicly available datasets and models in accordance with their applicable licenses and terms of use, and the work does not involve human subjects, private or sensitive data, or safety-critical applications.

\subsection*{Reproducibility Statement}

We provide the full RECAP algorithm and training objective in Section~\ref{sec:method} and Algorithm~\ref{alg:recap}. We report the model, dataset, optimization, sampling, and evaluation settings in Section~\ref{sec:experiments}. Additional implementation details, annotation procedures, and analysis protocols are documented in Appendix \ref{app:training-details}, including the semantic dependency annotation scheme, hyperparameters, and prompt templates.



\newpage
\bibliography{main}
\bibliographystyle{main}

\appendix
\section{Appendix}

\subsection{Related Work}
\label{app:related-work}
Token-efficient reasoning involves two distinct objectives: stopping once sufficient reasoning has been completed and reducing redundancy within the reasoning process itself. The former is addressed by inference-time or learned early-stopping methods~\citep{chen-etal-2026-step-grpo, sun-etal-2026-stop, xiang-etal-2026-thinking}. We focus on the latter objective that is to produce more compact reasoning traces without compromising solution quality~\citep{li-etal-2026-think-better}. Prompt-based methods encourage concise reasoning through explicit instructions, output constraints, or adaptive hints~\citep{10852493,lee2025how,tang2025concisehint}, but provide little guidance on which reasoning steps should be preserved. Post-training methods can provide more targeted control. Existing post-training methods for improving reasoning efficiency can be broadly grouped into length-regularized policy optimization, pruning and distillation, fine-grained credit assignment, and structure-aware attribution.

\paragraph{Length-Regularized Policy Optimization.} Length-regularized methods incorporate response length into rollout-level optimization, using fixed or adaptive penalties conditioned on correctness, problem difficulty, or training dynamics~\citep{NEURIPS2025_579b5b84, li2026drpo, peng2026think, li-etal-2026-leash}. While effective at shortening responses, these methods assign efficiency signals at the trajectory level and therefore cannot identify which individual reasoning steps are redundant.

\paragraph{Pruning and Distillation of Reasoning Traces.} Some methods construct concise training trajectories by pruning redundant content using teacher guidance, model-based importance estimates, or dependency analysis~\citep{jiang-etal-2026-drp, liu-etal-2026-pru, ma-etal-2026-reasoning, yuan-etal-2026-graph-based}. ThoughtFold further constructs sub-trajectories at different compression levels and uses masked preference optimization to encourage the model to bridge retained reasoning segments~\citep{liu2026thoughtfold}. These methods learn from preconstructed compressed trajectories rather than assigning step-specific credit within newly sampled rollouts, potentially limiting generalization to unseen redundancy patterns. They may still retain unnecessary steps or overcompress useful reasoning.

\paragraph{Fine-Grained Credit Assignment.} Fine-grained methods assign token- or step-level signals using answer likelihood, confidence, conditional mutual information, or attention~\citep{li-etal-2026-think-better, wang-etal-2026-stabilizing,he2026iapo,nie-etal-2026-attnpo}. BINGO estimates token significance and applies stronger efficiency pressure to less important tokens~\citep{liu-etal-2026-tokens}, while SwAP uses changes in correct-answer likelihood to redistribute excess-length penalties across steps~\citep{li2026stepwise}. Although more selective than trajectory-level objectives, these methods primarily estimate the utility of each unit in isolation and do not capture whether a step supports downstream deductions, provides context needed later, or merely repeats existing information. As a result, they may undercredit structurally important steps while overcrediting locally plausible but redundant ones.

\paragraph{Structure-Aware Attribution.} Several methods introduce tree- or graph-structured representations, primarily for efficient rollout sampling or finer-grained credit assignment rather than token efficiency. Tree-based methods organize trajectories through shared prefixes, while FlowTracer derives token-level credit from attention-induced information flow~\citep{ji2026tree,dong2026how}. GraphPO merges semantically equivalent states across rollouts and rewards shorter paths to shared states~\citep{zhan2026graphpo}. These methods capture shared rollout structure or model-internal information flow, rather than semantic dependencies among reasoning steps within a single derivation, and still rely on path length for token efficiency. RECAP instead directly constructs a semantic dependency graph over the reasoning steps of each rollout and uses it to attribute each step's contribution to the downstream derivation, allowing policy updates to reinforce useful steps more strongly while downweighting redundant ones.

\paragraph{Latent Reasoning.}
Latent-reasoning methods reduce explicit chain-of-thought generation by performing intermediate computation in continuous hidden states or recurrent depth~\citep{hao2025training, zhu2025scaling}. This changes the inference interface. Fewer generated tokens can correspond to additional latent model computation, making token counts not directly comparable with standard autoregressive reasoning. Moreover, existing approaches require specialized training or architectures rather than operating as drop-in post-training methods for the same reasoning model. We focus on methods that improve token efficiency under the standard autoregressive reasoning interface, where reductions in generated reasoning tokens are directly comparable.

\subsection{RECAP Algorithm}
\label{app:algorithm}
Algorithm~\ref{alg:recap} summarizes the complete RECAP training procedure, including dependency annotation, structural-responsibility propagation, step-efficacy computation, and advantage reshaping.

\begin{algorithm}[t]
\caption{RECAP Step-Level Credit Assignment}
\label{alg:recap}
\begin{algorithmic}[1]
\Require Prompt $q$; rollout group $\{\tau_k\}_{k=1}^{G}$; gold answer $a^\star=(a_1^\star,\ldots,a_T^\star)$; rollout-level advantages $\{A_k^{\mathrm{GRPO}}\}_{k=1}^{G}$; hyperparameters $\alpha,\beta,c,\epsilon$; policy model $\theta$; annotator model $\phi$
\Ensure Token-level advantages for all rollouts

\For{$k=1,\ldots,G$}
    \State Segment $\tau_k$ into semantic steps $\tau_k=(s_{k,1},\ldots,s_{k,N_k})$
    \State $\mathcal{G}_k \gets \textsc{BuildDependencyDAG}(\tau_k, \phi)$
    \State $\{\operatorname{resp}(s_{k,i})\}_{i=1}^{N_k}\gets \textsc{PropagateResponsibility}(\mathcal{G}_k)$)
    
    \State Compute the gold-answer log-likelihood before the first step:
    \vspace{-8pt}
    $$\mathcal{L}_{k,0}\gets\frac{1}{T}\sum_{t=1}^{T}\log \pi_{\theta}\left(a_t^\star\mid q,a_{<t}^\star \right)$$
    \vspace{-10pt}
    \For{$i=1,\ldots,N_k$}
        \State Compute the prefix-conditioned gold-answer log-likelihood:
        $$\mathcal{L}_{k,i}\gets\frac{1}{T}\sum_{t=1}^{T}\log \pi_{\theta}(a_t^\star\mid q,\tau_{k,\leq i},a_{<t}^\star)$$
        \State \(\Delta_{k,i}\gets\mathcal{L}_{k,i}-\mathcal{L}_{k,i-1}\)
        \State \(\widehat{\Delta}_{k,i}
        \gets\operatorname{clip}(\Delta_{k,i},-c,c)\)
        \State Compute the effective contribution score:
        $$ M_{k,i}\gets\operatorname{resp}(s_{k,i})\exp(\alpha\,\operatorname{sign}\left(A_k^{\mathrm{GRPO}}\right)\widehat{\Delta}_{k,i})$$
    \EndFor
    \State Let \(d_{k,i}\) be the number of tokens in \(s_{k,i}\), and compute
    $$D_k\gets\sum_{i=1}^{N_k}d_{k,i},\qquad\overline{M}_k\gets\frac{1}{D_k}\sum_{i=1}^{N_k}d_{k,i}M_{k,i}$$
    \For{\(i=1,\ldots,N_k\)}
        \State \(w_{k,i}\gets M_{k,i}/(\overline{M}_k+\epsilon)\)
        \State Compute the step-level advantage:
        $$ A_{k,i}^{\mathrm{RECAP}}\gets[(1-\beta)+\beta w_{k,i}]A_k^{\mathrm{GRPO}}$$
        \State Assign \(A_{k,i}^{\mathrm{RECAP}}\) to every token in \(s_{k,i}\)
    \EndFor
\EndFor
\State \Return token-level advantages
\end{algorithmic}
\end{algorithm}

\subsection{Annotation Quality}
\label{app:annotation_quality}

RECAP relies on an annotator to construct the dependency graph that determines its structural credit. 
Annotation errors may therefore distort the credit assigned to individual reasoning steps and ultimately affect policy optimization. To assess whether the structural signal is sufficiently reliable, we evaluate candidate annotators along two dimensions: annotation validity, which measures compliance with the required graph format, and annotation correctness, which measures the quality of the predicted dependencies. Because exact step dependencies lack ground-truth annotations, we estimate correctness through agreement with independent judges.

\paragraph{Annotation validity.}
We re-annotate 800 fixed rollouts generated by Qwen2.5-Math-1.5B-Instruct using Qwen3 models ranging from 0.6B to 32B, under both thinking and non-thinking settings, as well as Claude Haiku 4.5. We define the annotation validity rate as the fraction of outputs that can be parsed into fully typed, acyclic dependency graphs satisfying all format constraints.

\begin{figure}[t]
\begin{center}
\includegraphics[width=0.7\textwidth]{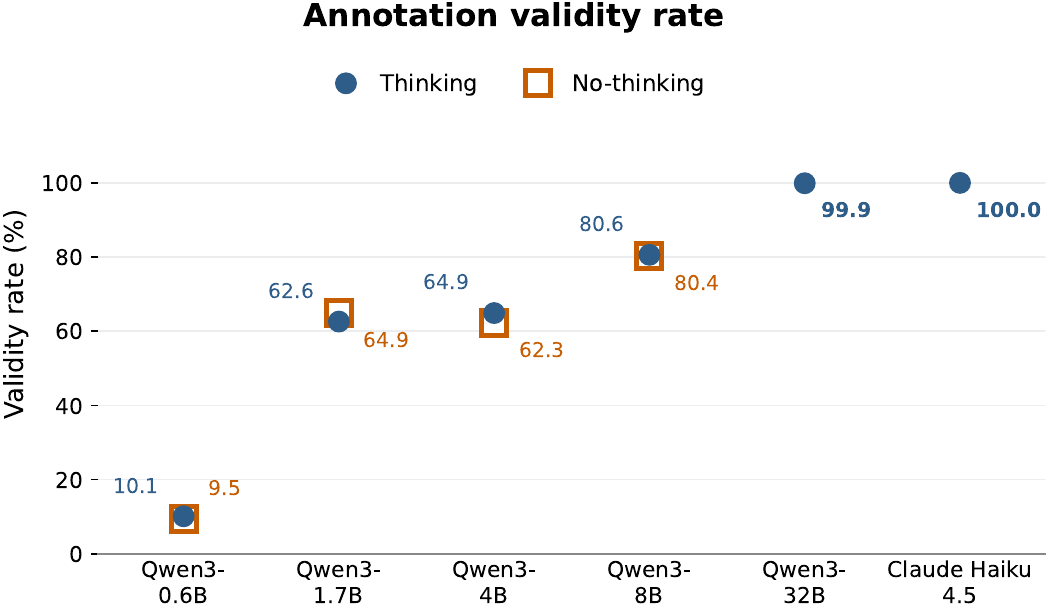}
\end{center}
\caption{\textbf{Annotation validity by annotator.} Fraction of rollouts whose dependency annotations parse into a valid, fully-typed, acyclic DAG, over $800$ frozen rollouts re-annotated by each of six candidate annotators. Tick marks on the Qwen3 bars show the thinking/no-thinking range ($\leq 2.5$ points). Only Qwen3-32B and Claude Haiku 4.5 achieve annotation validity rates above $99\%$.}
\label{fig:annotator-validity}
\end{figure}

Figure~\ref{fig:annotator-validity} that validity improves non-linearly with model scale. Among the Qwen3 models, the validity rate increases from $9.8\%$ at $0.6$B to $63.8\%$ at both $1.7$B and $4$B, $80.5\%$ at $8$B, and $99.9\%$ at $32$B. Switching between thinking and non-thinking modes changes validity by at most 2.5 percentage points at any model size. Claude Haiku 4.5 achieves a validity rate of $100.0\%$, comparable to Qwen3-32B. Since only Qwen3-32B and Claude Haiku 4.5 produce valid graphs nearly universally, we retain these two models for the subsequent annotation-correctness evaluation.

\paragraph{Annotation accuracy, Qwen3-32B vs. Claude Haiku 4.5}
\label{app:annotator-audit-acc}
Format compliance only verifies that an annotator produces a valid graph. It does not establish whether the predicted dependencies are correct. Therefore, we evaluate the two finalists from the validity study, Qwen3-32B and Claude Haiku 4.5, against the consensus of three independent judges (Claude Opus 4.8~\citep{anthropic2026claudetopus48}, GPT-5.5~\citep{openai2026gpt55}, and GLM-5~\citep{zeng2026glm}). These judges exhibit pairwise agreement rates of $75$--$85\%$ on this task, on 800 rollouts sampled from the DAPO training set, comprising 16 fixed rollouts for each of 50 prompts.

We evaluate step-to-step support, context, and restatement edges, as well as dependencies from reasoning steps to the final answer. We report pooled edge accuracy and edge-type F1 scores. Pooled accuracy is the fraction of decidable edge candidates for which the annotator's binary prediction agrees with the judge consensus, aggregated across all dependency types. The aggregation weights each type by its frequency. Here, a decidable cell is one where the judges reach consensus and the test model itself parses the relevant step successfully. Here, an edge candidate is considered decidable when the judges reach consensus and the candidate annotator successfully parses the relevant steps. F1 score for each dependency type, is the harmonic mean of their precision and recall, where precision measures the fraction of predicted edges supported by the judge consensus and recall measures the fraction of consensus edges recovered by the annotator.
Table~\ref{tab:annotator-accuracy} shows that Claude Haiku 4.5 outperforms Qwen3-32B on every metric, with a 35-point advantage in pooled accuracy. 

The validity and agreement evaluations show that Claude Haiku 4.5 matches Qwen3-32B in format reliability while agreeing substantially more often with the independent judge panel. We therefore use Claude Haiku 4.5 as the annotator for RECAP.

\begin{table}[t]
\caption{\textbf{Claude Haiku 4.5 agrees with the judge panel far more than Qwen3-32B, on the same rollouts.} Pooled accuracy and per-type F1 against the $\geq 2$-of-$3$ judge consensus (Claude Opus 4.8, GPT-5.5, GLM-5), $800$ rollouts ($50$ prompts $\times$ $16$ rollouts).}
\label{tab:annotator-accuracy}
\begin{center}
\begin{tabular}{lcccc}
\toprule
Model & Pooled accuracy \% & Support F1 & Context F1 & Restate F1 \\
\midrule
Qwen3-32B   & $41.9$          & $62.8$          & $13.3$          & $12.3$          \\
Claude Haiku 4.5   & $\mathbf{77.0}$ & $\mathbf{83.0}$ & $\mathbf{35.4}$ & $\mathbf{51.3}$ \\
\bottomrule
\end{tabular}
\end{center}
\end{table}

\begin{table*}[t]
\centering
\small
\caption{Ablation of the annotator models under fixed hyperparameter settings: $\beta=0.3, \alpha=1, \gamma_{context}=0.5, \tau_{\Delta}=2, \tau_w=5$.}
\label{tab:annotator-swap}
\begin{tabular}{@{}lcccccccc@{}}
\toprule
\multirow{2}{*}{Annotator} & \multicolumn{2}{c}{GSM8K} & \multicolumn{2}{c}{MATH-500} & \multicolumn{2}{c}{AIME 2024} & \multicolumn{2}{c}{AIME 2025} \\
\cmidrule(lr){2-3} \cmidrule(lr){4-5} \cmidrule(lr){6-7} \cmidrule(lr){8-9}
 & pass@1$\uparrow$ & tokens$\downarrow$ & pass@1$\uparrow$ & tokens$\downarrow$ & pass@1$\uparrow$ & tokens$\downarrow$ & pass@1$\uparrow$ & tokens$\downarrow$ \\
\midrule
\textbf{Haiku 4.5} & \textbf{84.6} & \textbf{237} & \textbf{73.2} & \textbf{509} & \textbf{25.8} & \textbf{1401} & \textbf{11.2} & \textbf{1135} \\
Qwen3-32B & 82.2 & 259 & 69.2 & 512 & 24.6 & 1421 & 8.7 & 1222 \\
\bottomrule
\end{tabular}
\end{table*}

\paragraph{Effect of annotator choice.}
Beyond annotation accuracy, we further examine whether the choice of annotator affects downstream training outcomes with all other hyperparameters fixed. Note that this setting was not fully tuned and differs from the one used in Table~\ref{tab:main}.
 As shown in Table~\ref{tab:annotator-swap}, RECAP with Haiku 4.5 achieves higher $pass@1$ and fewer tokens than Qwen3-32B on all four benchmarks. This suggests that annotation quality translates into more effective credit assignment and supports our use of Haiku 4.5 in the main experiments. Since we do not retune hyperparameters for each annotator, the comparison reflects performance under a common configuration rather than each annotator's best achievable result.

\begin{table*}[t]
\centering
\small
\setlength{\tabcolsep}{3pt}
\caption{Annotator models on the same four benchmarks, scored with the identical prompt, answer extractor and grader as the trained-checkpoint tables.(p@1=pass@1, p@8=pass@8)}
\label{tab:api-models-sampled}
\begin{tabular}{@{}lcccccccccccc@{}}
\toprule
Model & \multicolumn{3}{c}{GSM8K} & \multicolumn{3}{c}{MATH-500} & \multicolumn{3}{c}{AIME 2024} & \multicolumn{3}{c}{AIME 2025} \\
\cmidrule(lr){2-4} \cmidrule(lr){5-7} \cmidrule(lr){8-10} \cmidrule(lr){11-13}
 & p@1$\uparrow$ & p@8$\uparrow$ & tok$\downarrow$ & p@1$\uparrow$ & p@8$\uparrow$ & tok$\downarrow$ & p@1$\uparrow$ & p@8$\uparrow$ & tok$\downarrow$ & p@1$\uparrow$ & p@8$\uparrow$ & tok$\downarrow$ \\
\midrule
Haiku 4.5 & \textbf{95.8}\,\tiny{$\pm$0.3} & \textbf{98.0} & 473 & \textbf{92.6}\,\tiny{$\pm$0.5} & \textbf{96.2} & 1594 & \textbf{65.8}\,\tiny{$\pm$3.0} & \textbf{83.3} & 3909 & \textbf{46.7}\,\tiny{$\pm$5.4} & \textbf{70.0} & 4096 \\
Qwen3-32B & 94.3\,\tiny{$\pm$0.4} & 97.7 & \textbf{273} & 82.1\,\tiny{$\pm$0.7} & 92.4 & \textbf{834} & 30.0\,\tiny{$\pm$4.7} & 46.7 & \textbf{2601} & 19.2\,\tiny{$\pm$6.4} & 40.0 & \textbf{2563} \\
\bottomrule
\end{tabular}
\end{table*}

\paragraph{Reasoning capability of annotator models.}
We also evaluate the annotator models themselves on the same four reasoning benchmarks, as shown in Table~\ref{tab:api-models-sampled}. Haiku 4.5 consistently achieves higher $pass@1$ and $pass@8$ than Qwen3-32B, with particularly large gaps on MATH-500 and AIME. This suggests stronger mathematical reasoning capability for interpreting the target-domain traces. Since dependency annotation is answer-blind and provides neither gold answers nor solution trajectories, this capability is used to assess relations among policy-generated steps rather than to provide task-level supervision.

\subsection{Training Details.} 
\label{app:training-details}
\subsubsection{Setup}
During training, we set the sampling temperature to $1.0$ and sample $n=16$ rollouts per prompt, with a maximum response length of $8192$ tokens. We use a training batch size of $128$ prompts and a mini-batch size of $1024$ rollouts, with two policy-update epochs per training step, and train for four passes over the prompt pool. Filtering prompts longer than $1024$ tokens removes exactly one of the $2048$ training prompts, leaving the pool just below $16$ full batches. With \texttt{drop\_last}, this results in $15$ training steps per epoch and $60$ training steps in total, with the remaining $127$ prompts discarded at the end of each epoch. The actor is optimized with AdamW using a constant learning rate of $1\times10^{-6}$ without warmup, gradient clipping at $1.0$, a PPO clip ratio of $0.2$, and no entropy bonus. Following GRPO, the KL regularizer is applied directly to the policy loss rather than incorporated into the reward. We use the low-variance $k_3$ estimator with coefficient $10^{-3}$ against a frozen reference policy. Rollouts are generated with vLLM using dynamic batching, and a single random seed ($42$) is used for the dataloader, vLLM engine, and sampler.

Advantage shaping introduces no additional optimizer hyperparameters. RECAP modifies only the advantage-construction stage, so the optimizer, rollout configuration, and all other training settings are kept identical to those of the corresponding GRPO and controlled baselines. For structural responsibility, only the relative ratios among the edge-type coefficients matter during propagation. We therefore fix $\gamma_{\mathrm{support}}=1$ and $\gamma_{\mathrm{restate}}=0$, and tune only $\gamma_{\mathrm{context}}$. Support edges form the main dependency backbone and receive full edge weight, whereas restatement edges introduce no new reasoning contribution and therefore receive zero weight. 
We tune $\gamma_{\mathrm{context}}$ separately for each model on the validation set. Table~\ref{tab:hyper-gamma} shows the hyperparameter study for DeepSeek-R1-Distill-Qwen-7B as an example. The selected values are $\gamma_{\mathrm{context}}=0.5$ for Qwen2.5-Math-7B 
and $\gamma_{\mathrm{context}}=0.75$ for DeepSeek-R1-Distill-Qwen-7B.

The remaining hyperparameters control the strength and stability of step-level reshaping. We clip step efficacy at $|\Delta|\le\tau_\Delta$, with model-specific thresholds chosen from the offline efficacy statistics in Table~\ref{tab:delta-distribution}. Qwen2.5-Math-7B typically produces fewer than $20$ semantic steps per trace, so we set $\tau_\Delta=2$, approximately the 90th percentile of its offline distribution. Given the typical trace length, this clips only about two steps per trace. DeepSeek-R1-Distill-Qwen-7B produces substantially longer traces. After merging its traces to a target of at most $50$ semantic steps, we set $\tau_\Delta=4$, which is near the 95th percentile of its offline distributions, again limiting clipping to only a small number of steps per trace ($\sim 2\text{ steps}$). We further clip the normalized step weight at $w\le\tau_w$, using $\tau_w=5$ for Qwen2.5-Math-7B and $\tau_w=10$ for DeepSeek-R1-Distill-Qwen-7B. These thresholds are chosen near the 99th percentile of the corresponding offline $w$ distributions in Table~\ref{tab:w-distribution}, preventing a small number of extreme weights from dominating the update while leaving the vast majority of steps unaffected.

The efficacy scaling coefficient $\alpha$ and residual-mixing coefficient $\beta$ are selected through hyperparameter search on the validation set. We use $\alpha=0.5$ for Qwen2.5-Math-7B and $\alpha=1.0$ for DeepSeek-R1-Distill-Qwen-7B, with $\beta=0.3$ for both models. In each case, we select the configuration with the highest validation accuracy. For brevity, Tables~\ref{tab:hyper-alpha} and~\ref{tab:hyper-beta} report the sensitivity to $\alpha$ and $\beta$ on Qwen2.5-Math-7B. DeepSeek-R1-Distill-Qwen-7B is tuned using the same validation protocol.

We evaluate the policy every $15$ training steps on the held-out $512$-prompt DAPO-Math validation split using greedy decoding. As with the training set, we filter out prompts longer than 1024 tokens, and this removed two of the 512 validation prompts. The reported checkpoint is selected by the highest validation $pass@1$ and is frozen before any test-set evaluation. Training rollouts are sampled at temperature $1.0$ to maintain sufficient within-group diversity for informative relative advantages.

For the annotator, we set $temperature=0$, $maxTokens=4096$, and $enable\_thinking=False$.

The test datasets and their sizes are reported in Appendix Table~\ref{tab:test-size}.

\begin{table}[t]
\caption{The test dataset sizes.}
\label{tab:test-size}
\begin{center}
\begin{tabular}{lcccc}
\toprule
Dataset & GSM8K & MATH-500 & AIME 2024 & AIME 2025 \\
\midrule
Size  & 1319 & 500  & 30 & 30\\
\bottomrule
\end{tabular}
\end{center}
\end{table}

\begin{table}[t]
\caption{\textbf{Hyperparameter Study of $\gamma_{context}:\gamma_{support}$} on DeepSeek-R1-Distill-Qwen-7B. We fixed $\gamma_{support}=1$ and adjust $\gamma_{context}$ to change its ratio. }
\label{tab:hyper-gamma}
\begin{center}
\begin{tabular}{lcccc}
\toprule
$\gamma_{context}$ & \textbf{0.33} & \textbf{0.50} & \textbf{0.75} & \textbf{1.00} \\
\midrule
Accuracy  & 0.6725 & 0.6451  & 0.6804 & 0.6804\\
\bottomrule
\end{tabular}
\end{center}
\end{table}

\begin{table}[t]
\centering
\small
\setlength{\tabcolsep}{4pt}
\caption{Offline distribution of the signed step efficacy $\Delta$ on the two backbones, measured with the production scorer over 400 rollouts per model. 
$\Delta$ is a per-step log-probability difference in nats per gold token, i.e.\ an \emph{absolute} quantity whose scale is model-dependent. neg. denotes the fraction of steps with negative efficacy.
}
\label{tab:delta-distribution}

\textbf{
$\Delta$ distribution}\\[2pt]
\begin{tabular}{@{}lrrrrrrrrrr@{}}
\toprule
& & \multicolumn{5}{c}{$|\Delta|$ quantiles} & \multicolumn{2}{c}{signed $\Delta$} & \multicolumn{2}{c}{clip} \\
\cmidrule(lr){3-7} \cmidrule(lr){8-9} \cmidrule(lr){10-11}
Backbone & steps/roll. & med & p90 & p95 & p99 & max & mean & neg.($\%$) & $\tau_\Delta$ & \% \\
\midrule
Qwen2.5-Math-7B & 18.5 & 0.363 & 1.996 & 2.909 & 5.590  & 13.9 & $-0.022$ & 51.5 & 2.0 & 10.0 \\
R1-Distill-7B   & 36.5 & 0.602 & 2.893 & 3.988 & 6.519  & 14.2 & $+0.023$ & 49.5 & 4.0 & 5.0 \\
\bottomrule
\end{tabular}
\end{table}

\begin{table}[ht]
\centering
\small
\setlength{\tabcolsep}{5pt}
\caption{Distribution of the per-step multiplier $M$ and of the normalised weight $w=\mathrm{clip}\!\left(\frac{M}{\bar{M}+\epsilon},\,0,\,\tau_w\right)$.
$w_{\mathrm{raw}}$ is the same quantity before the $\tau_w$ clip, so the gap between $w$ and $w_{\mathrm{raw}}$ at p99 and max is exactly what clipping removes. 
}
\label{tab:w-distribution}
\begin{tabular}{@{}lcc@{}}
\toprule
& Qwen2.5-Math-7B & R1-Distill-7B  \\
\midrule
$\tau_\Delta$ / $\tau_w$ / $\alpha$ & 2.0 / 5 / 0.5 & 4.0 / 10 / 1.0  \\
$\beta$ & 0.3 & 0.3 \\
\midrule
$resp$ median & 1.0000 & 0.3869 \\
$resp$ zero fraction & 12.9\% & \textbf{27.0\%} \\
$|\Delta|$ median (in training) & 0.4928 & 0.5817 \\
$M$ median & 0.8361 & 0.2462 \\
\midrule
$w$ median & \textbf{0.7757} & 0.1195 \\
$w$ mean & 0.8955 & 0.6572 \\
$w$ p90 & 1.8057 & 1.5245 \\
$w$ p95 & 2.3394 & 3.1872 \\
$w$ p99 & 3.9964 & 9.7834 \\
$w$ max & 5.0000 & 10.0000 \\
\midrule
$w_{\mathrm{raw}}$ p90 & 1.81 & 1.52 \\
$w_{\mathrm{raw}}$ p95 & 2.34 & 3.19 \\
$w_{\mathrm{raw}}$ p99 & 4.00 & 9.96 \\
$w_{\mathrm{raw}}$ max & \textbf{161.8} & 61.4 \\
\midrule
median per-rollout $\max w$ & 1.533 & 6.373 \\
token-weighted mean $w$ & \textbf{1.0000} & 0.7490 \\
$\tau_w$ trigger rate & \textbf{0.59\%} & 0.99\% \\
$\tau_\Delta$ trigger rate & \textbf{16.83\%} & 5.25\%  \\
\bottomrule
\end{tabular}
\end{table}

\begin{table}[ht]
\caption{\textbf{Hyperparameter Study of $\alpha$} under fixed hyperparameters: $\beta=0.3, \gamma_{context}=0.5$.}
\label{tab:hyper-alpha}
\begin{center}
\begin{tabular}{lcccc}
\toprule
$\alpha $ & \textbf{0.3} & \textbf{0.5} & \textbf{0.8} & \textbf{1.0} \\
\midrule
Accuracy  &0.4510 & 0.5000 & 0.4667 & 0.4824\\
\bottomrule
\end{tabular}
\end{center}
\end{table}

\begin{table}[ht]
\caption{\textbf{Hyperparameter Study of $\beta$} under fixed hyperparameters: $\alpha=1, \gamma_{context}=0.5$.}
\label{tab:hyper-beta}
\begin{center}
\begin{tabular}{lccccc}
\toprule
$\beta $ & \textbf{0} & \textbf{0.2} & \textbf{0.3} & \textbf{0.5} &\textbf{1.0}\\
\midrule
Accuracy & $0.4490$   & $0.4804$ & $ 0.4824$ & $0.4510$ & 0.4431 \\
\bottomrule
\end{tabular}
\end{center}
\end{table}

\subsubsection{Step splitting}
\label{app:step-target}
In our experiments, each rollout is split into steps at blank lines ($\backslash n\backslash n$). For Qwen2.5-Math-7B, this raw $\backslash n\backslash n$ split is used as is: it yields ${\sim}16$ steps of
${\sim}77$ tokens each, a granularity that matches what a step
intuitively means. The two long-thinking models break this assumption.
On $128$ rollouts generated at training sampling parameters, DeepSeek-R1-Distill-Qwen-7B's raw split yields a median of $174$ fragments per rollout at a median of only $22$ tokens each ($65.1\%$ are $\leq 30$ tokens). The maximum number of the fragments in a rollout can reach $447$. Since the RECAP annotator emits ${\sim}18$ output tokens of dependency declarations per step, a $174$-step trace exhausts its output budget ($4096$), its dependency graph
fails to parse (${\sim}29\%$), and the rollout contributes \emph{no} structural signal at all. 

We address this by introducing a per-rollout \emph{step target}. Rollouts at or
below the target keep the raw split and rollouts above it have adjacent short
steps merged until they fit. We use two experiments set a propriate target: an offline screen over many targets to find where annotation stops being reliable, and training runs at a coarse grid to choose among the targets the screen leaves undecided.

\paragraph{Experiment 1: Offline Annotation Screen}

\begin{figure}[t]
\begin{center}
\includegraphics[width=0.6\textwidth]{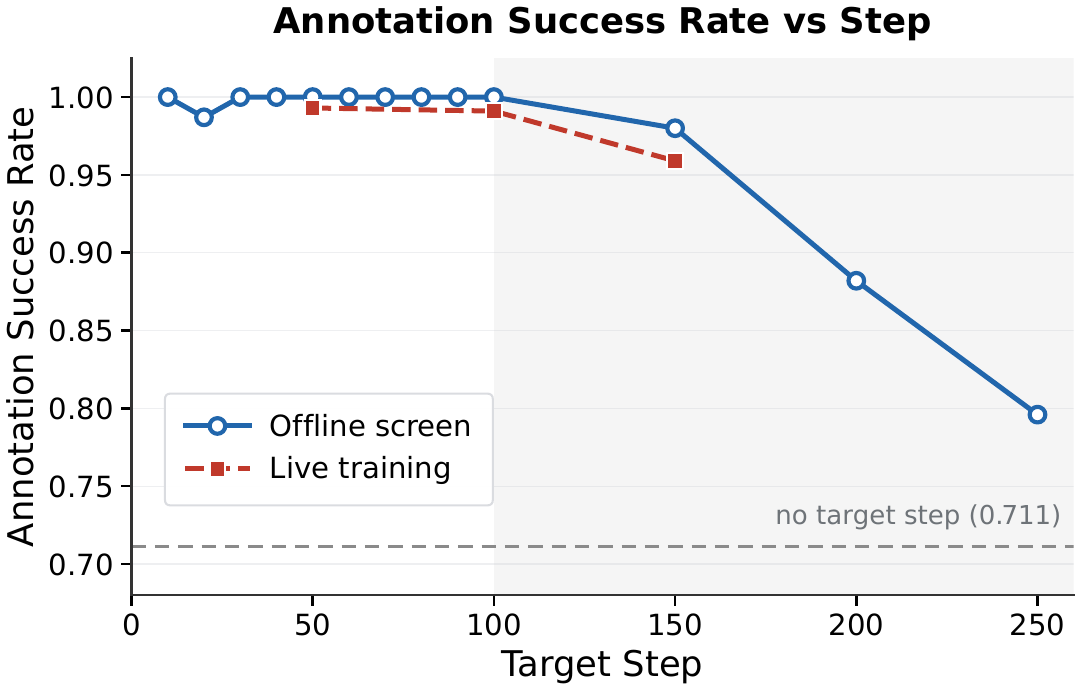}
\end{center}
\caption{\textbf{Annotation reliability bounds the step target from above.}
whose dependency annotation is successful (all steps are annotated) versus the
per-rollout step target.
\emph{Offline screen}: $128$ frozen R1-Distill-7B rollouts re-annotated by
Haiku~4.5, $3$ passes per setting. 
\emph{Live training}: the same metric during the three runs of
the training (\texttt{MAX\_STEPS\_PER\_ROLLOUT}$=50,100,150$, $512$ rollouts
$\times$ 4 training steps).
Dashed line: the raw split ($0.711$).
Reliability saturates at $100$ and is still $0.980$ at $150$.}
\label{fig:okfrac-steps}
\end{figure}

We use Claude Haiku~4.5 to annotate the $128$ fixed rollouts processed with different target step. The same $128$ frozen rollouts areannotated at $14$ step targets by Haiku~4.5 ($4096$ output tokens, $3$ passes each), scored over the $76$ rollouts containing \verb|\boxed| (a rollout truncated before its answer makes the annotator refuse regardless of the target). Figure~\ref{fig:okfrac-steps} shows that the annotation reliability is flat at $1.000$ for targets $\leq 100$, \textbf{still $0.980$ at $150$}, then decays ($0.882$ at $200$, $0.796$ at $250$, $0.711$ with the raw split). We then search the target steps below 150 steps because they are ${\sim}100\%$ \emph{ok\_frac}. The screen bounds the target from above but cannot discriminate within $[50, 150]$, since annotation is reliable throughout that range. Specifically, we did a widely spaced targets search with small training ($512$ rollouts
$\times$ 4 training steps): $150$ (just past saturation), $100$ ($86$ tok/step, closest to Qwen's $77$), and $50$ (the $2\times$-coarser control).

\begin{table}[t]
\caption{\textbf{Training runs at the three step targets}
(DeepSeek-R1-Distill-Qwen-7B, four steps, validation every two steps).}
\label{tab:step-target}
\begin{center}
\begin{tabular}{lccc}
\toprule
 & \textbf{target $50$} & \textbf{target $100$} & \textbf{target $150$} \\
\midrule
Offline annotation success & $1.000$ & $1.000$ & $0.980$ \\
Offline tokens / step & $155$ & $86$ & $73$ \\
\midrule
Task accuracy & $\mathbf{0.614}$ & $0.596$ & $0.590$ \\
\midrule
Live annotation success & $\mathbf{0.992}$--$\mathbf{0.994}$
                        & $0.986$--$0.996$
                        & $0.953$--$0.963$ \\
Live mean tokens / step & $152$ & $87$ & $73$ \\
Training Time (s/step) & $\mathbf{461}$--$\mathbf{644}$ & $491$--$677$ & $495$--$836$ \\
\bottomrule
\end{tabular}
\end{center}
\end{table}

\paragraph{Experiment 2: Training at the three targets}

Each target was used for a four-step training ($\beta = 0.3$, $\gamma_{\mathrm{context}} = 0.5$, $512$ rollouts per step, validation on \texttt{dapo\_val512} every two steps). Table~\ref{tab:step-target} reports the outcome trained on DeepSeek-R1-Distill-Qwen-7B. Target $50$ wins on validation accuracy and holds the highest live annotation reliability, never falling below $0.992$, against $150$'s drop to $0.953$--$0.963$. It also leads to a faster training speed ($461 s$ -- $644 s$ per step). Based on the results, we adopt $50$ as the step target (threshold) of for thinking models.

\subsubsection{Training Time Comparison}
\label{app:training-time}

Table~\ref{tab:training-time} reports the approximate wall-clock training time of RECAP and the post-training baselines under the available hardware configurations. All runs use $8$ GPUs, $60$ training steps, and a maximum response length of $8192$ tokens. Since not every method was run on both H100 and A100 GPUs, we report measured times only without cross-hardware normalization.

\begin{figure*}[ht]
    \centering
    \begin{subfigure}[t]{0.48\textwidth}
        \centering
        \includegraphics[width=\linewidth]{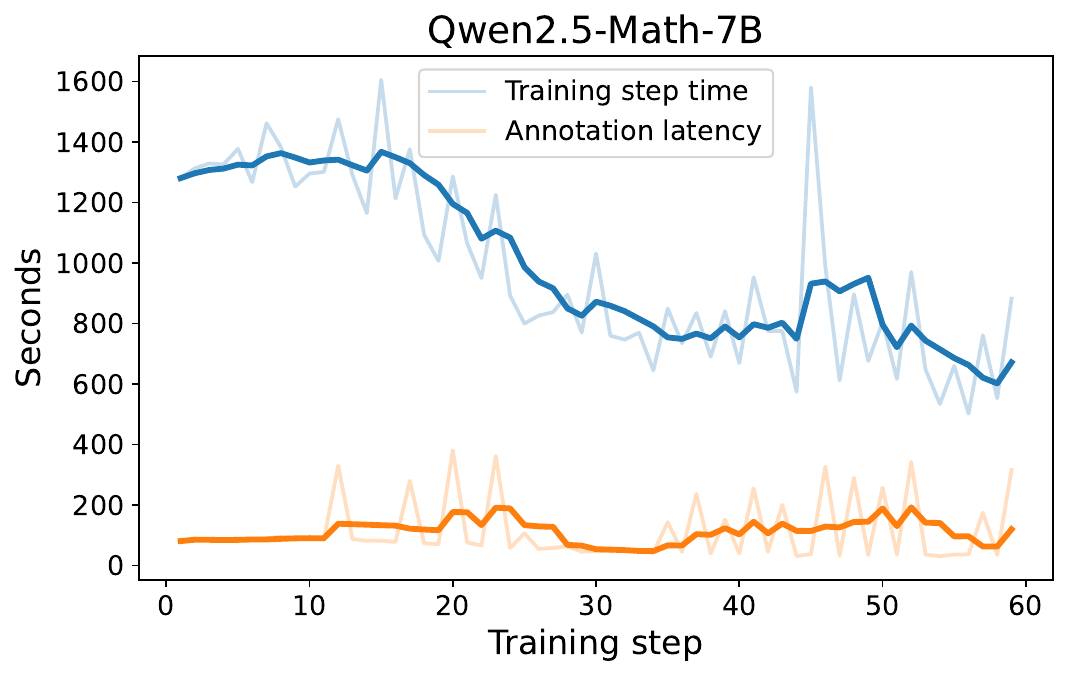}
        \caption{Qwen2.5-Math-7B.}
    \end{subfigure}
    \hfill
    \begin{subfigure}[t]{0.48\textwidth}
        \centering
        \includegraphics[width=\linewidth]{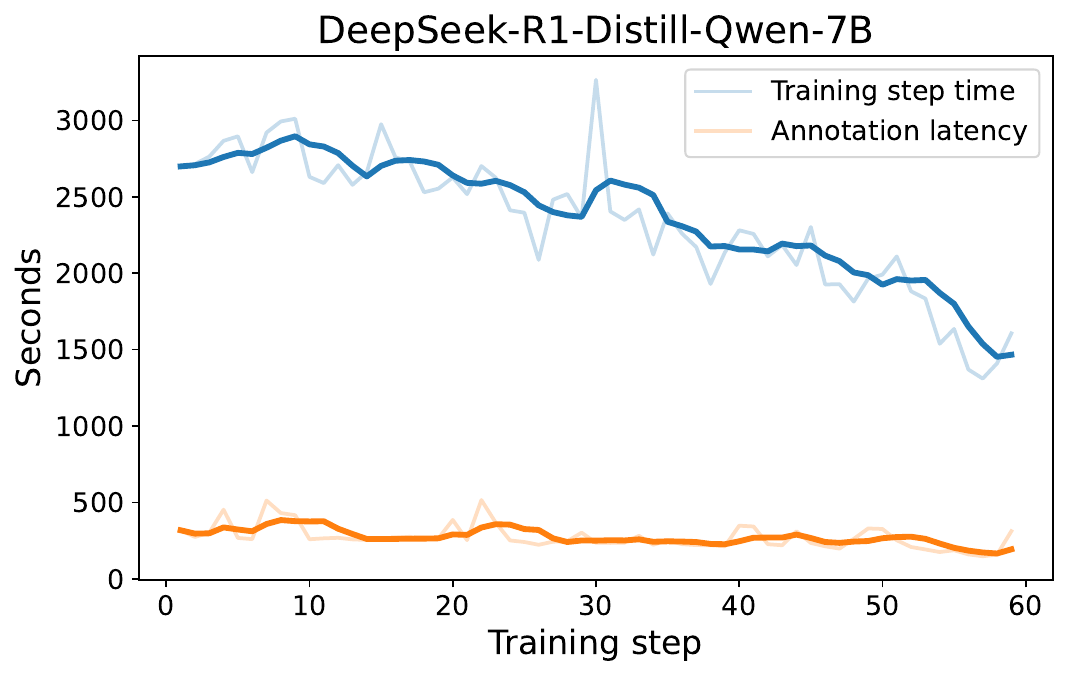}
        \caption{DeepSeek-R1-Distill-Qwen-7B.}
    \end{subfigure}
    \caption{\textbf{Per-step training time and annotation latency of RECAP on \textbf{$8\times$ A100}.} Light curves show raw measurements and dark curves show 5-step rolling averages. Annotation latency measures the wall-clock time spent waiting for dependency annotations.}
    \label{fig:training-time-breakdown}
\end{figure*}

Figure~\ref{fig:training-time-breakdown} further decomposes RECAP's per-step training time. Dependency annotation accounts for only about $12\%$ of the recorded step time for both models. Annotation latency occasionally spikes due to external API throttling, but these spikes do not dominate the overall training cost. On 8$\times$H100, standard GRPO takes $\sim300$s per step on average, while RECAP takes $\sim510$s with Qwen2.5-Math-7B. The additional overhead of RECAP consists of $\sim149$s per step for computing step efficacy and $\sim61$s for annotating semantic dependencies. With DeepSeek-R1-Distill-Qwen-7B, whose rollouts are relatively longer, standard GRPO takes $\sim912$s per step on average, while RECAP takes $\sim1320$s. In this setting, RECAP incurs an additional $\sim250$s per step for computing step efficacy and $\sim158$s for semantic dependency annotation.

\begin{table}[ht]
\centering
\small
\caption{\textbf{Approximate wall-clock training time} on $8\times$ H100 or $8\times$ A100 GPUs. ``--'' denotes configurations that were not measured.}
\label{tab:training-time}
\begin{tblr}{
  colspec   = {lcccccccc},
  colsep    = 4pt,
  rowsep    = 0.8pt,
  column{1} = {leftsep  = 0pt},
  column{9} = {rightsep = 0pt},
}
\toprule
\textbf{Method} & \textbf{$8\times$ H100} & \textbf{$8\times$ A100} \\
\midrule
\SetRow{bg=panelbg, abovesep=2pt, belowsep=2pt} \SetCell[c=3]{c} \textit{Qwen2.5-Math-7B}\\
GRPO                & 5h        & --        \\
Simple Length       & 4h 40m    & --        \\
DDCA                & 5h 20m    & --        \\
LEASH               & 3h 35m    & --        \\
BINGO               & --        & 6h         \\
ThoughtFold         & 10h       & --        \\
RECAP (ours)        & 8h 30m    & 16h 23m   \\

\midrule
\SetRow{bg=panelbg, abovesep=2pt, belowsep=2pt} \SetCell[c=3]{c} \textit{DeepSeek-R1-Distill-Qwen-7B}\\
GRPO                & 15h 12m   & --        \\
Simple Length       & --        & 19h 55m   \\
DDCA                & --        & 32h 20m   \\
LEASH               & --        & 26h       \\
BINGO               & --        & 11h 40m   \\
ThoughtFold         & --        & 32h 10m   \\
RECAP (ours)        & 22h       & 38h 41m   \\

\bottomrule
\end{tblr}
\end{table}

\begin{table*}[t]
\centering
\small
\setlength{\tabcolsep}{4pt}
\caption{\textbf{Dependency-DAG Analysis of Dead-End Reasoning.} Structural properties of the dependency DAGs for Qwen2.5-Math-7B rollouts over the full test sets of all four benchmarks, measured on the \emph{same} $n=8$ samples that produced the accuracy tables. \emph{ML} = share of rollouts with more than one sink in the step sub-graph, i.e.\ containing work whose result no later step uses and that never reaches the answer. \emph{ISO} = share with at least one step of total degree zero, connected to nothing at all. \textbf{Bold} marks the best column value. }
\label{tab:dag-topology}
\begin{tabular}{@{}lcrrrrrr@{}}
\toprule
 & mean & \multicolumn{2}{c}{Haiku 4.5$^{\dagger}$} & \multicolumn{2}{c}{Opus 4.8} & \multicolumn{2}{c}{GPT-5.5} \\
\cmidrule(lr){3-4} \cmidrule(lr){5-6} \cmidrule(lr){7-8}
Model & steps & ML$\downarrow$ & ISO$\downarrow$ & ML$\downarrow$ & ISO$\downarrow$ & ML$\downarrow$ & ISO$\downarrow$ \\
\midrule
Base & 11.7 & 43.2 & 26.0 & 43.3 & 38.0 & 42.7 & 34.3 \\
Prompt-guided  & 10.0 & 37.2 & 22.3 & 37.7 & 31.5 & 37.9 & 30.6 \\
GRPO & 8.2 & 71.9 & 60.0 & 76.6 & 69.9 & 78.2 & 74.5 \\
\midrule
Simple Length & 6.9 & 37.5 & 20.0 & 40.3 & 30.0 & 38.8 & 31.2 \\
DDCA & 9.3 & 46.8 & 13.3 & 52.4 & 34.1 & 50.2 & 35.7 \\
LEASH & 8.9 & 44.7 & 11.8 & 48.5 & 29.6 & 47.0 & 33.0 \\
Bingo & 9.3 & 44.9 & 20.5 & 47.9 & 33.1 & 45.5 & 34.3 \\
ThoughtFold & 8.4 & 43.7 & 11.7 & 49.7 & 32.8 & 49.2 & 36.0 \\
\midrule
\textbf{RECAP (ours)} & \textbf{4.0} & \textbf{11.0} & \textbf{5.9} & \textbf{10.8} & \textbf{7.0} & \textbf{15.6} & \textbf{12.8} \\
\bottomrule
\end{tabular}
\end{table*}

\subsubsection{Training Dynamics Analysis.}
\label{app:training-dynamics}

\begin{figure}[t]
\begin{center}
\includegraphics[width=1\textwidth]{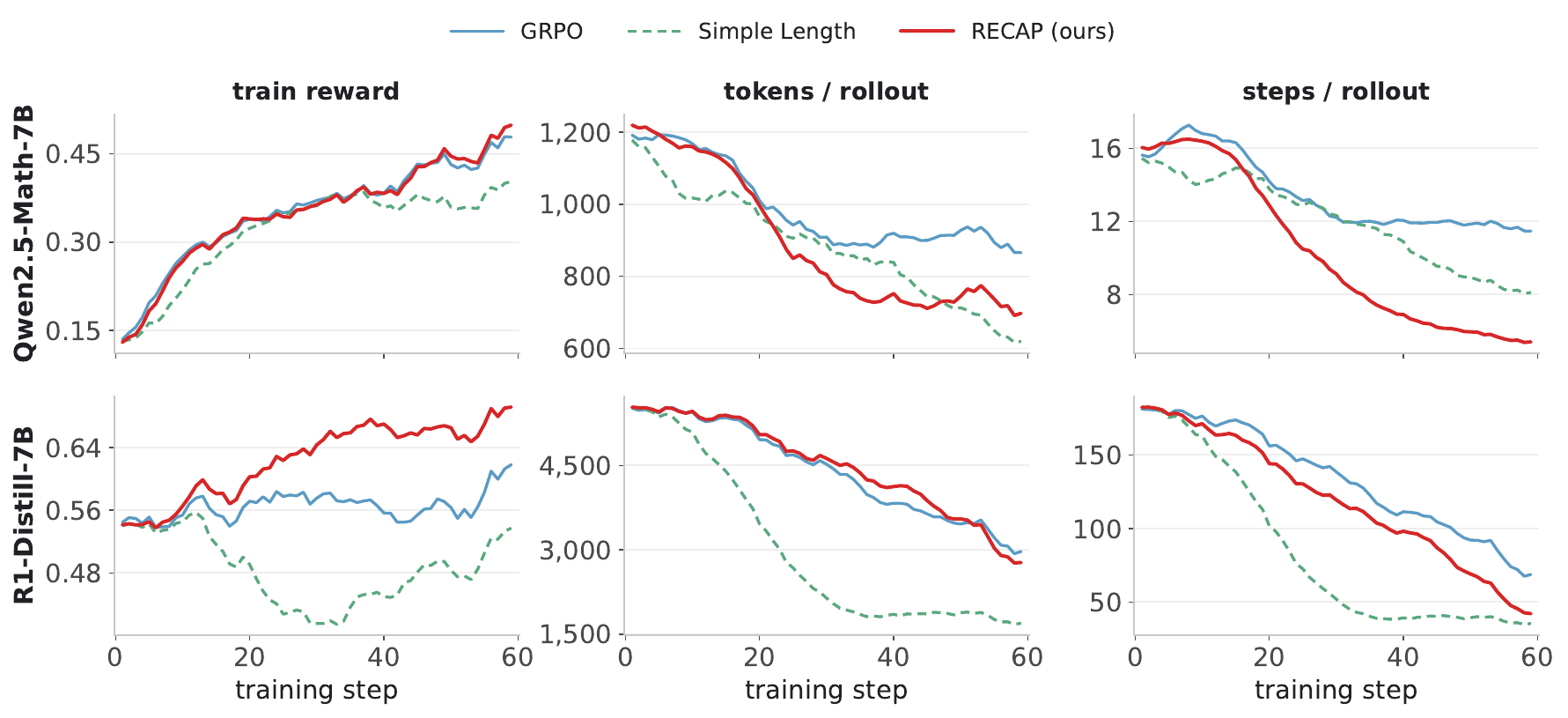}
\end{center}
\caption{Training dynamics.}
\label{fig:train_dyn}
\vspace{-15pt}
\end{figure}

\paragraph{Reward and reasoning length during training.}
Figure~\ref{fig:train_dyn} compares GRPO, Simple Length, and RECAP throughout training. We focus on these three methods because they isolate the main mechanisms of interest: standard outcome-based RL without an explicit efficiency objective, trajectory-level length regularization, and RECAP's step-level redundancy-aware credit assignment.

Across both models, RECAP progressively reduces response length and the number of reasoning steps while maintaining strong training reward. Relative to GRPO, RECAP ends training with substantially fewer tokens and steps while achieving higher reward. Simple Length often reduces token count more aggressively, but at substantially lower reward, especially on DeepSeek-R1-Distill-Qwen-7B. These trajectories shows that reducing length alone is easy, whereas RECAP aims to remove reasoning while preserving the computation needed for successful solutions.

\begin{figure}[t]
\begin{center}
\includegraphics[width=0.5\textwidth]{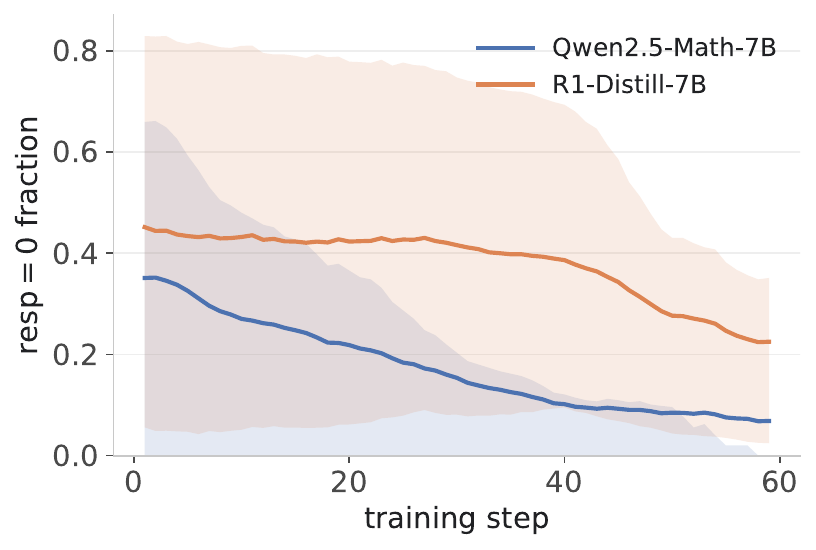}
\end{center}
\vspace{-15pt}
\caption{Fraction of reasoning steps that cannot reach the final answer ($resp=0$) during RECAP training: mean over each step's $\sim$2000 rollouts, band = inter-quartile range across rollouts, both smoothed over 5 steps. }
\label{fig:evo_of_dead_end_steps}
\end{figure}

\paragraph{Structural redundancy decreases during training.}
Figure~\ref{fig:evo_of_dead_end_steps} tracks the fraction of reasoning steps that receive zero structural responsibility under RECAP. This fraction decreases throughout training for both models. Thus, as training progresses, a larger fraction of generated steps participates in the credited downstream derivation. Together with the simultaneous reduction in total step count in Figure~\ref{fig:train_dyn}, this trend is consistent with RECAP progressively suppressing structurally unused reasoning rather than merely shortening individual steps. It also provides training-time evidence complementary to the test-time dead-end analysis in Table~\ref{tab:dag-topology} and Section~\ref{sec:rq3}.

\subsubsection{Atomic Reasoning Operation Annotation}
\label{app:atomic_op_def}
To distinguish a reduction in reasoning content from a change in how reasoning is segmented or expressed, we manually annotate atomic reasoning operations on a subset of training rollouts. This analysis is independent of the semantic-step boundaries used by RECAP: annotators inspect the reasoning trace itself and count the underlying operations, regardless of whether several operations appear in the same semantic step or one operation spans multiple textual fragments.

We define an \emph{atomic reasoning operation} as a minimal action that changes the current solution state or introduces new information that is subsequently used in the derivation. The details are shown in Table~\ref{tab:atomic-op}.

\begin{table}[t]
\centering
\small
\setlength{\tabcolsep}{4pt}
\caption{\textbf{Annotation rubric for atomic reasoning operations.}
An atomic reasoning operation is a minimal action that changes the current solution state or introduces new information used in the derivation. Pure planning statements, restatements, exact repetitions, and off-task text are excluded. Incorrect operations are still counted.}
\label{tab:atomic-ops}
\begin{tabular}{@{}p{0.18\linewidth}p{0.43\linewidth}p{0.31\linewidth}@{}}
\toprule
\textbf{Operation} & \textbf{Definition} & \textbf{Example} \\
\midrule
Setup / notation
& Introduces a variable, object, or representation that is subsequently used.
& $BD=x,\; DC=6-x$ \\

Fact recall
& Invokes a mathematical fact or named theorem needed by the derivation.
& Vieta's formulas; angle-bisector theorem \\

Equation derivation
& Derives a new mathematical relation from existing information.
& $x^2-14x+48\leq0 \Rightarrow (x-6)(x-8)\leq0$ \\

Substitution
& Substitutes an established value or relation into another expression.
& Substitute $b=6c-3$ into the third equation. \\

Numerical computation
& Performs an arithmetic or numerical calculation.
& $1503\times4=6012$ \\

Case analysis
& Evaluates one distinct branch of an enumeration or case split.
& Analyze the case of four-digit even numbers. \\

Verification
& Checks a previously derived result, analytically or computationally.
& Verify $\binom{23}{4}+\binom{23}{5}=\binom{24}{5}$. \\

Conclusion
& Converts the established derivation into the requested result.
& Conclude that the answer is $x=6$. \\
\bottomrule
\end{tabular}
\label{tab:atomic-op}
\vspace{-15pt}
\end{table}

\subsubsection{Prompts}
\begin{figure}[t]
\begin{center}
\includegraphics[width=0.9\textwidth]{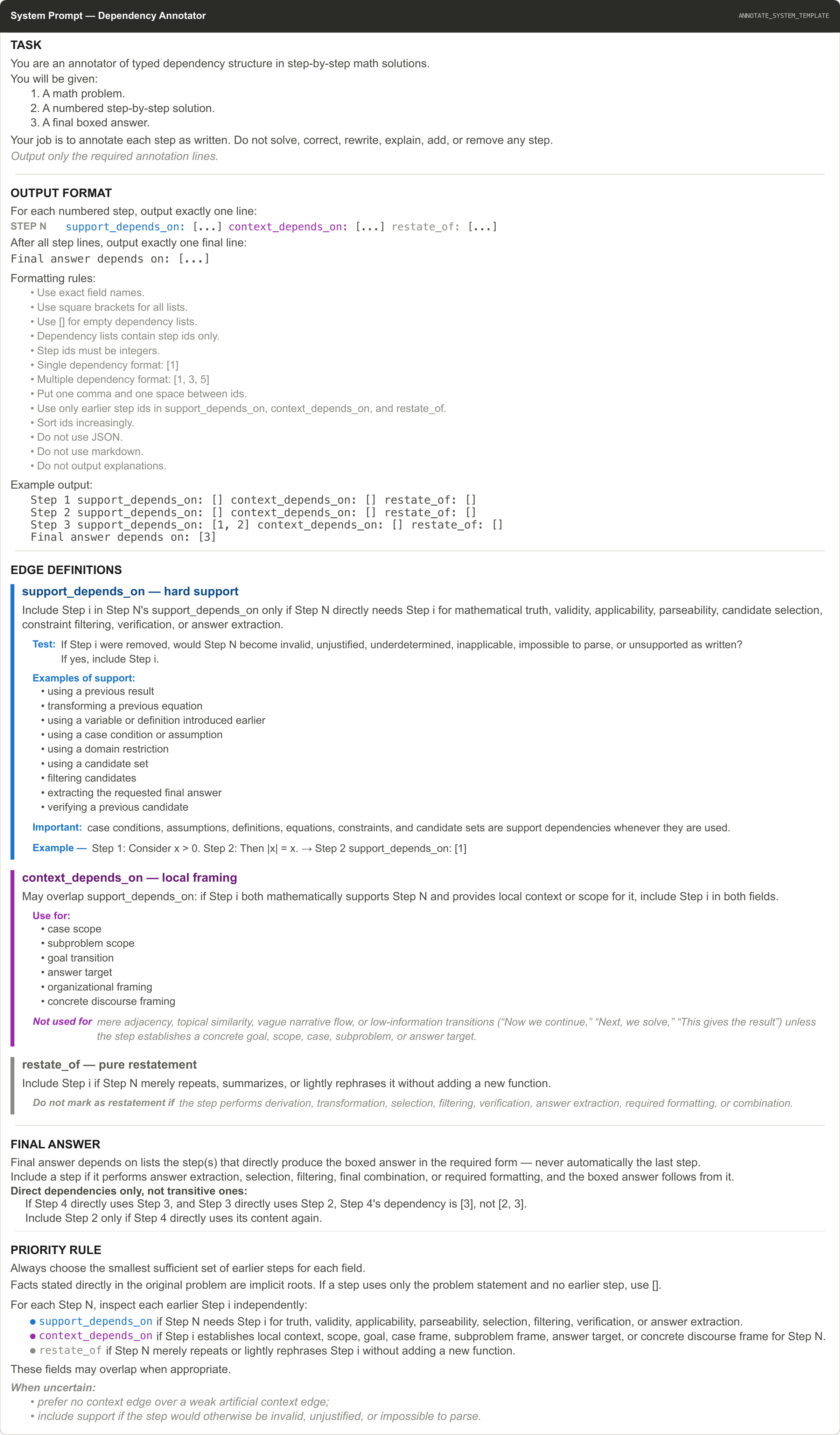}
\end{center}
\caption{System prompt given to the annotator.}
\label{fig:system-prompt-card}
\end{figure}

\begin{figure}[t]
\begin{center}
\includegraphics[width=0.9\textwidth]{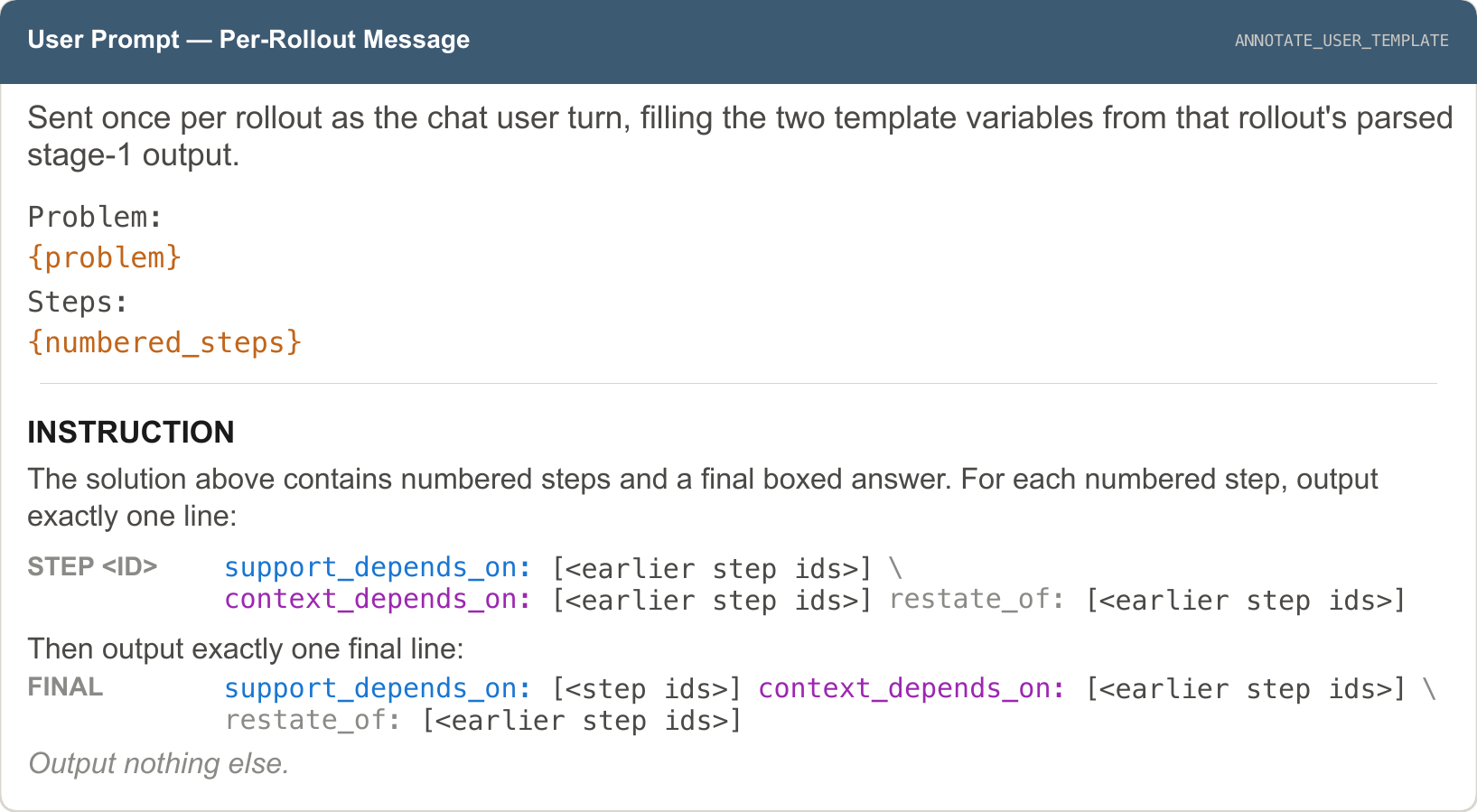}
\end{center}
\caption{User prompt given to the annotator}
\label{fig:user-prompt-card}
\end{figure}

\paragraph{Annotator Prompts}
\label{app:annotator-prompt}
Figures~\ref{fig:system-prompt-card} and~\ref{fig:user-prompt-card} show the two prompts for the annotator models.

\begin{figure}[t]
\begin{center}
\includegraphics[width=0.9\textwidth]{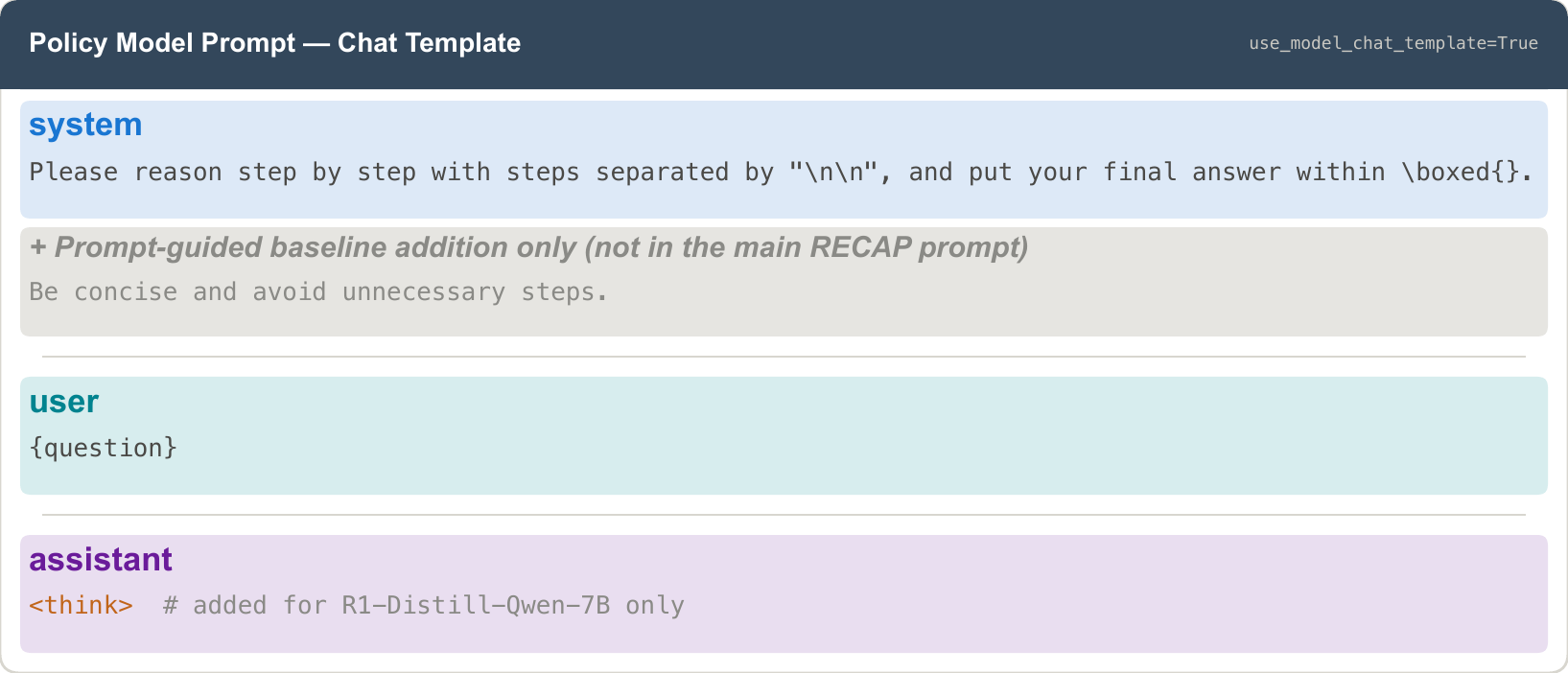}
\end{center}
\caption{\textbf{Policy model prompt (chat-template mode).}  System (blue) is the unified instruction that replaces the default template. The gray band directly below is the Prompt-guided-baseline-only addition. User (teal) is the unmodified question. Assistant (violet) is where generation begins. \texttt{<think>} is used for R1-Distill-Qwen-7B only, to enable its thinking capability. 
Qwen2.5-Math-7B does not have explicit thinking mode.}
\label{fig:policy-prompt-card}
\end{figure}

\paragraph{Policy Model Prompt}
\label{app:policy-prompt}
Figure~\ref{fig:policy-prompt-card} shows the prompt the \emph{policy} model itself is trained and evaluated on
(\texttt{use\_model\_chat\_template=True}). To enable R1-Distill-Qwen-7B's thinking mode, we add \texttt{<think>} to its chat template only. Qwen2.5-Math-7B does not have the explicit thinking mode. 

\end{document}